%% file: main.tex
\documentclass[sigconf,nonacm]{acmart}
\usepackage{layout}
\usepackage{hyperref}

\usepackage{textcomp}
\usepackage{xcolor}
\usepackage[caption=false,font=footnotesize]{subfig}
\usepackage{algorithm}
\usepackage{algorithmic}
\newcommand{\our}{PAT}

\newcommand{\para}[1]{\noindent \textbf{#1 }}
\newcommand\blfootnote[1]{%
  \begingroup
  \renewcommand\thefootnote{}%
  \footnotetext{#1}%
  \endgroup
}
\AtBeginDocument{%
  }

\copyrightyear{2018}
\acmYear{2018}
\acmDOI{XXXXXXX.XXXXXXX}
\acmConference[Conference acronym 'XX]{Make sure to enter the correct
  conference title from your rights confirmation email}{June 03--05,
  2018}{Woodstock, NY}
\acmISBN{978-1-4503-XXXX-X/2018/06}

\begin{document}

%%
%% The "title" command has an optional parameter,
%% allowing the author to define a "short title" to be used in page headers.
\title{Accelerating Long-Tail Generation in Synchronous RLHF Training via Adaptive Tensor Parallelism}

%%
%% The "author" command and its associated commands are used to define
%% the authors and their affiliations.
%% Of note is the shared affiliation of the first two authors, and the
%% "authornote" and "authornotemark" commands
%% used to denote shared contribution to the research.
% \author{Ben Trovato}
% \authornote{Both authors contributed equally to this research.}
% \email{trovato@corporation.com}
% \orcid{1234-5678-9012}
% \author{G.K.M. Tobin}
% \authornotemark[1]
% \email{webmaster@marysville-ohio.com}
% \affiliation{%
%   \institution{Institute for Clarity in Documentation}
%   \city{Dublin}
%   \state{Ohio}
%   \country{USA}
% }
\author{
Long Zhao$^{\dagger,\S}$ \;
Qinghe Wang$^{\dagger,\S}$ \;
Jiaan Zhu$^{\mathparagraph}$ \;
Youhui Bai$^{\mathparagraph}$ \;
Zewen Jin$^{\mathparagraph}$ \;
Chaoyi Ruan$^{\ddagger}$ \newline
Shengnan Wang$^{*}$ \;
Cheng Li$^{\mathparagraph,\S}$
\\
{\normalsize
$^{\dagger}$ Anhui University \;
$^{\mathparagraph}$ University of Science and Technology of China \;
$^{\ddagger}$ National University of Singapore \;
$^{*}$ Independent Researcher \newline
$^{\S}$ Institute of Artificial Intelligence, Hefei Comprehensive National Science Center
}
}

%%
%% By default, the full list of authors will be used in the page
%% headers. Often, this list is too long, and will overlap
%% other information printed in the page headers. This command allows
%% the author to define a more concise list
%% of authors' names for this purpose.
\renewcommand{\shortauthors}{Trovato et al.}

%%
%% The abstract is a short summary of the work to be presented in the
%% article.
% \begin{abstract}

% \end{abstract}
\input{0.abstract}
%%
%% The code below is generated by the tool at http://dl.acm.org/ccs.cfm.
%% Please copy and paste the code instead of the example below.
%%
% \begin{CCSXML}
% <ccs2012>
%  <concept>
%   <concept_id>00000000.0000000.0000000</concept_id>
%   <concept_desc>Do Not Use This Code, Generate the Correct Terms for Your Paper</concept_desc>
%   <concept_significance>500</concept_significance>
%  </concept>
%  <concept>
%   <concept_id>00000000.00000000.00000000</concept_id>
%   <concept_desc>Do Not Use This Code, Generate the Correct Terms for Your Paper</concept_desc>
%   <concept_significance>300</concept_significance>
%  </concept>
%  <concept>
%   <concept_id>00000000.00000000.00000000</concept_id>
%   <concept_desc>Do Not Use This Code, Generate the Correct Terms for Your Paper</concept_desc>
%   <concept_significance>100</concept_significance>
%  </concept>
%  <concept>
%   <concept_id>00000000.00000000.00000000</concept_id>
%   <concept_desc>Do Not Use This Code, Generate the Correct Terms for Your Paper</concept_desc>
%   <concept_significance>100</concept_significance>
%  </concept>
% </ccs2012>
% \end{CCSXML}

% \ccsdesc[500]{Do Not Use This Code~Generate the Correct Terms for Your Paper}
% \ccsdesc[300]{Do Not Use This Code~Generate the Correct Terms for Your Paper}
% \ccsdesc{Do Not Use This Code~Generate the Correct Terms for Your Paper}
% \ccsdesc[100]{Do Not Use This Code~Generate the Correct Terms for Your Paper}

%%
%% Keywords. The author(s) should pick words that accurately describe
%% the work being presented. Separate the keywords with commas.
%% A "teaser" image appears between the author and affiliation
%% information and the body of the document, and typically spans the
%% page.

% \received{20 February 2007}
% \received[revised]{12 March 2009}
% \received[accepted]{5 June 2009}

%%
%% This command processes the author and affiliation and title
%% information and builds the first part of the formatted document.
\maketitle

\blfootnote{Long Zhao and Qinghe Wang equally contributed to this work.}
\input{1.intro}
\input{2.background}
\input{3.observation}
\input{4.design}
\input{5.evaluation}

\input{6.relatedwork}

\bibliographystyle{ACM-Reference-Format}
\bibliography{ref}

\end{document}

%% file: 0.abstract.tex
\begin{abstract}
Reinforcement Learning from Human Feedback (RLHF) has become a key post-training paradigm for improving model quality. However, the synchronous three-stage RLHF pipeline is often bottlenecked by the generation stage, where response-length skew causes the effective batch size to shrink rapidly during decoding, leaving GPUs underutilized while a few long responses remain unfinished. Mainstream frameworks employ a static tensor parallelism (TP) configuration that cannot adapt to changing batch characteristics, leaving substantial performance headroom unexplored.

We propose \our{}, an adaptive TP method that dynamically reconfigures TP during the generation stage of each RLHF iteration. \our{} introduces two key techniques. First, a predictor-guided online reconfiguration method decides both the reconfiguration point and the target TP configuration based on offline profiling, triggering reconfiguration only when the predicted latency benefit outweighs the reconfiguration overhead. Second, a lightweight online reconfiguration mechanism updates only the states and layouts affected by TP changes: it adapts unfinished decoding states through a cost-model-based choice between KV-cache migration and recomputation, performs in-place weight resharding, and reuses cached communication groups. We implement \our{} on top of SGLang and integrate it with the VeRL framework. Evaluations on LLaMA3.1-8B and Qwen3-14B using DeepScaleR show that \our{} reduces generation latency by up to 34.6\% and end-to-end RLHF training iteration latency by up to 27.2\% compared to the original VeRL setup.

\end{abstract}
\keywords{RLHF, tensor parallelism, parallelism switching}

%% file: 1.intro.tex
\section{Introduction}
Reinforcement Learning from Human Feedback (RLHF)~\cite{bai2022training} has emerged as an effective post-training paradigm for aligning model outputs with human preferences and substantially improving the capabilities of large language models. Several recent prominent models, for example, DeepSeek-R1~\cite{guo2025deepseek}, Kimi K2~\cite{team2025kimi}, and OpenAI o3~\cite{el2025competitive}, have adopted or explored RLHF training strategies to enhance their model quality.

The RLHF training pipeline proceeds \textit{synchronously} through three stages in each iteration. First, the \textbf{generation stage} produces responses from the target LLM for given prompts. Second, the \textbf{preparation stage} evaluates generated responses, computing rewards or other auxiliary signals. Finally, the \textbf{training stage} consumes the outputs of the preparation stage, computes the training loss using the prepared signals, and updates the target LLM.

%%CL Mainstream RLHF frameworks, such as VeRL~\cite{sheng2025hybridflow}, adopt a colocation strategy, deploying all the aforementioned stages on the same GPU servers, thereby avoiding additional data movement and synchronization overhead across stages. In this situation, sample generation becomes the primary performance bottleneck, mainly due to the pronounced long-tail length distribution of the generated samples~\cite{zhong2025optimizing}. Specifically, as illustrated in the upper part of Fig.~\ref{switch-example}, the generation stage naturally evolves from an aligned phase to a tail phase. The aligned phase represents the initial stage in which most samples have comparable response lengths and are completed nearly simultaneously. In contrast, the tail phase represents the later stage in which most samples are finished, except for a few long-tail samples. As a result, in the tail phase, GPU resources are severely underutilized, and the subsequent preparation and training stages cannot start until all tail samples finish. In our benchmark with a 16K maximum response length and a batch size of 128, the average per-GPU achieved throughput drops from 13.71 TFLOPS in the aligned phase to 0.11 TFLOPS in the tail phase.

\begin{figure}[!t]
\centerline{\includegraphics[width=1\linewidth]{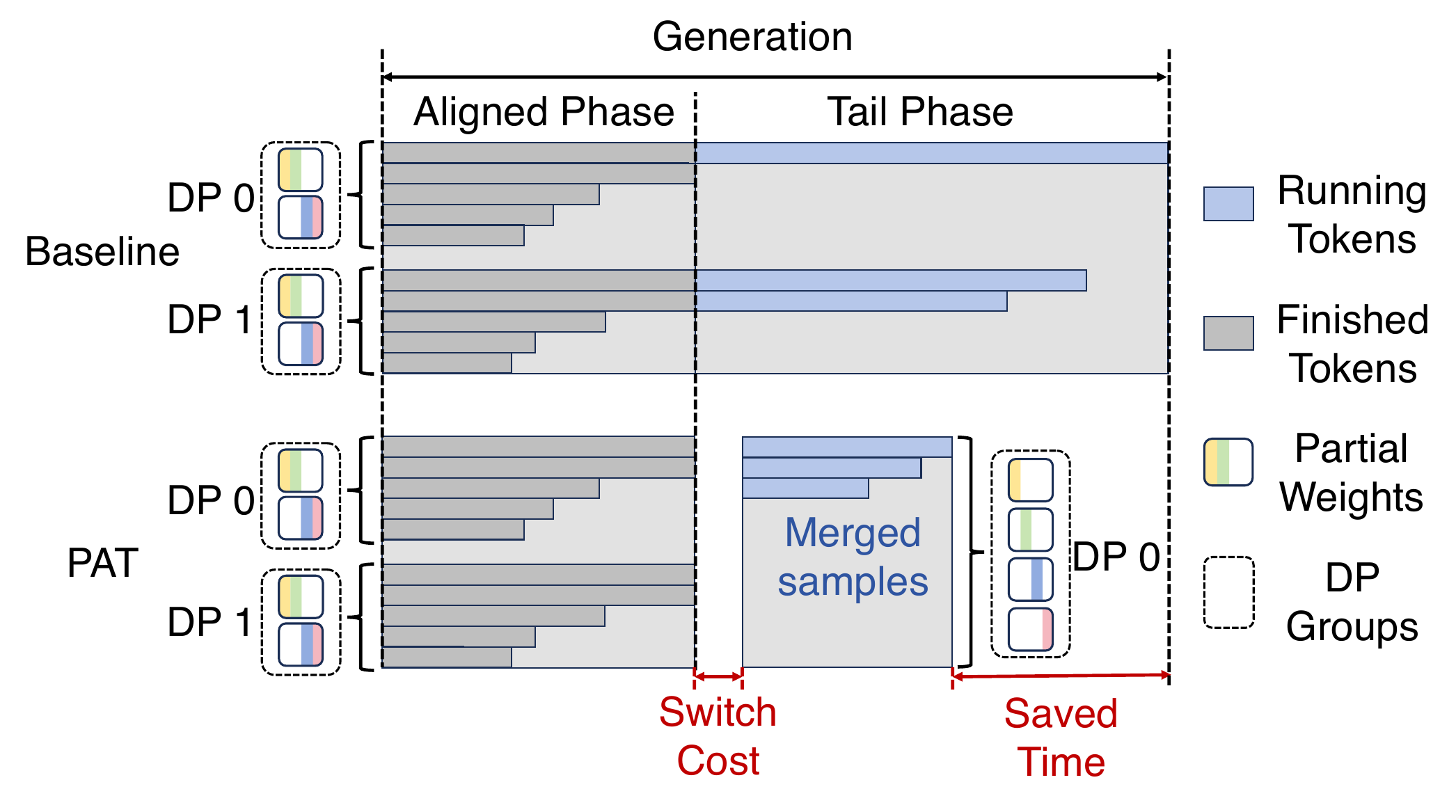}}
\caption{Example of TP/DP reconfiguration during generation. \our{} switches from $(TP2,DP2)$ in the aligned phase to $(TP4,DP1)$ in the tail phase.}
\label{switch-example}
\end{figure}

Mainstream RLHF frameworks, such as VeRL~\cite{sheng2025hybridflow}, prefer to colocate all three stages on the same GPU servers. Combined with synchronous execution, this reduces data movement and synchronization overhead across stages. Nevertheless, sample generation remains the primary performance bottleneck. Fig.~\ref{switch-example} illustrates the root cause of this inefficiency. Initially, most responses have similar lengths and finish nearly simultaneously—a phase we term the \textbf{aligned phase}. Later, a small number of exceptionally long responses continue executing, entering what we call the \textbf{tail phase}. The GPU resource consumption across the two phases is highly imbalanced, leaving GPU resources severely underutilized during the tail phase. In our benchmark with a maximum response length of 16K tokens and a batch size of 128, the average per-GPU throughput drops from 13.71 TFLOPS in the aligned phase to only 0.11 TFLOPS in the tail phase. Thus, accelerating RLHF crucially depends on accelerating the tail phase caused by response-length skew.

A closer investigation reveals that these frameworks typically employ a static tensor parallelism (TP) configuration throughout the entire generation stage. While the TP degree plays a key role in determining generation performance, it remains fixed regardless of the changing batch characteristics. Under a fixed GPU budget, the large effective batch size in the aligned phase favors a smaller TP degree to achieve higher throughput. In contrast, the few remaining samples in the tail phase benefit from a larger TP degree, which reduces decoding latency and improves GPU utilization. This observation motivates adapting the TP degree on the fly, specifically increasing it as the effective batch size shrinks.

However, dynamically reconfiguring TP during generation is not straightforward, due to several challenges. First, TP reconfiguration incurs non-trivial overhead. Specifically, unfinished samples must be transferred or recomputed under the new parallelism layout, model weights must be resharded across the new TP group, and communication groups must be consistently updated. 

Second, the benefit of reconfiguration involves a trade-off between the remaining tail workload and the reconfiguration overhead. Reconfiguring too early hurts aligned-phase throughput, as higher TP introduces extra communication and reduces data parallelism (DP) concurrency\footnote{TP is often combined with DP in practice. Although we primarily change the TP degree, doing so also results in changes to the DP configuration.}. Reconfiguring too late leaves insufficient tail work to amortize the reconfiguration overhead. Consequently, an effective system must decide when reconfiguration is beneficial and perform it with low overhead.

To address these challenges, we propose \our{}, an adaptive tensor parallelism framework that performs TP reconfiguration within the generation stage of each RLHF iteration.
As shown in the lower part of Fig.~\ref{switch-example}, \our{} keeps a throughput-oriented low-TP/high-DP configuration in the aligned phase, where many samples are still active. It then switches to a latency-oriented high-TP/low-DP configuration in the tail phase only when the predicted benefits outweigh the reconfiguration overhead. 
\our{} builds on two key techniques:

First, \our{} devises a predictor-guided online switching method to determine both the switching point and the target TP configuration. 
Relying on offline-profiled data, the predictor estimates the remaining generation latency under each candidate configuration and incorporates the one-time reconfiguration overhead into the decision. 
\our{} triggers reconfiguration only when the predicted latency benefit on the remaining tail samples can amortize the switching cost.

Second, \our{} makes reconfiguration lightweight through online runtime state and layout adaptation. 
Instead of reinitializing the inference engine under the target TP configuration, \our{} updates only the TP-dependent states and layouts. 
For unfinished samples, it reconstructs decoding states through a cost-model-based choice between KV cache migration and recomputation. 
For model weights, it performs in-place resharding to match the target TP layout. 
It also reuses cached communication groups across iterations to avoid repeated group construction. 
Together, these mechanisms turn TP reconfiguration into a low-overhead operation, allowing tail samples to resume decoding efficiently under the new configuration.

We build \our{} on SGLang~\cite{zheng2024sglang}, a widely used inference engine for RLHF generation, and integrate it with VeRL~\cite{sheng2025hybridflow}, one of the leading RLHF frameworks, enabling easy adoption. We evaluate \our{} by training LLaMA3.1-8B~\cite{dubey2024llama} and Qwen3-14B~\cite{yang2025qwen3} on DeepScaleR~\cite{luo2025deepscaler}. Compared with the original VeRL setup, \our{} reduces generation latency by up to 34.6\%, yielding up to 27.2\% lower RLHF training iteration latency.

\if 0
Our contributions are summarized as follows:

\begin{itemize}
    \item We analyze the long-tail phenomenon in synchronous RL scenarios and explore the potential benefits of adaptive TP reconfiguration.
    \item We propose \our{}, an efficient TP switching system that leverages idle GPU resources to accelerate long-tail request generation.
    \item We conduct experiments comparing the state-of-the-art RL framework VeRL on the LLaMA3.1-8B and LongWriter-6k datasets.
\end{itemize}
\fi 

%% file: 2.background.tex
\section{Background and Motivation}

\subsection{Workflow of RLHF}
\begin{figure}[!t]
\centering
\includegraphics[width=0.8\linewidth]{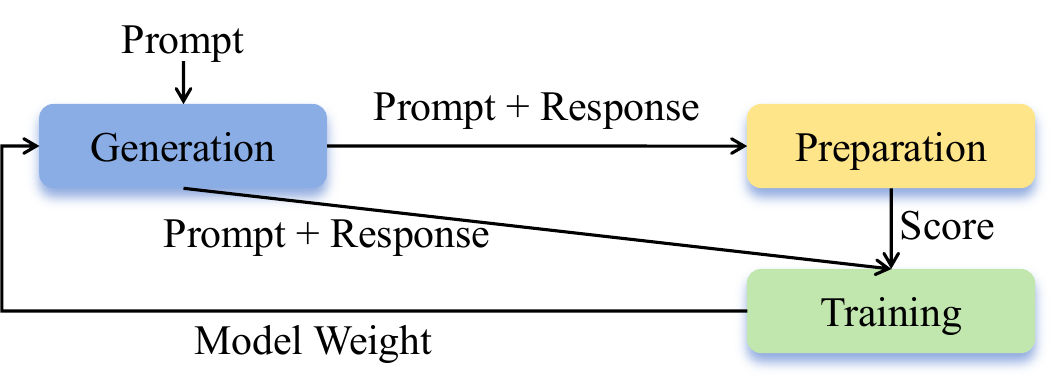}
\caption{Typical workflow of the iterative RLHF training.}
\label{fig:rlhf-workflow}
\end{figure}

% RLHF is a common approach for aligning LLMs with human preferences~\cite{ouyang2022training, guo2025deepseek}. It typically involves three steps: supervised fine-tuning, reward model training, and reinforcement learning (e.g., PPO~\cite{schulman2017proximal}, GRPO~\cite{shao2024deepseekmath}). Our work focuses on the reinforcement learning step, where iterative generation, scoring, and model updating dominate the overall training cost. As shown in Fig.~\ref{fig:rlhf-workflow}, it can be further decomposed into the following three stages\footnote{Although there are multiple algorithmic variants of RL (e.g., PPO~\cite{schulman2017proximal}, GRPO~\cite{shao2024deepseekmath}, DAPO~\cite{yu2025dapo}), their overall workflows appear similar.}.
RLHF is a common approach for aligning LLMs with human preferences~\cite{ouyang2022training, guo2025deepseek}. 
It typically consists of supervised fine-tuning, reward model training, and reinforcement learning. 
This work focuses on the reinforcement learning step, including algorithms such as PPO~\cite{schulman2017proximal}, GRPO~\cite{shao2024deepseekmath}, and DAPO~\cite{yu2025dapo}, where iterative generation, scoring, and model updates dominate the overall training cost. 
As shown in Fig.~\ref{fig:rlhf-workflow}, this step can be further decomposed into three stages.

% \noindent\textbf{Generation stage.} Given a batch of prompts, the target LLM to be trained (i.e., the actor model) generates corresponding responses in an auto-regressive manner. This stage consists of the prefill phase and the decoding phase.
% The prefill computes all input prompts in parallel, generates the first token for each, and constructs the corresponding KV cache.
% The decoding generates output tokens autoregressively, reusing the KV states in the KV cache to efficiently produce each subsequent token, which is commonly memory-bandwidth-bound and typically requires a large batch size to achieve high GPU utilization~\cite{agrawal2024taming}. 
\noindent\textbf{Generation stage.}
Given a batch of prompts, the actor model generates responses in an
autoregressive manner. This stage includes prefill, which processes the
input prompts and builds the KV cache, and decoding, which reuses KV
states to generate subsequent tokens. Decoding is typically
memory-bandwidth-bound and benefits from large effective batches~\cite{agrawal2024taming}.

\noindent\textbf{Preparation stage.}
Given the generated prompt-response pairs, auxiliary models or modules
such as reward and reference models compute rewards, log probabilities,
or other training signals used for loss computation.
% After obtaining the prompt–response pairs, a single forward pass is performed through multiple models: the critic model produces values, the reference model computes reference log probabilities, and the reward model provides rewards. Unlike the prefill stage, this step requires only one forward computation without generating any token.

\noindent\textbf{Training stage.}
The actor model performs forward and backward passes on the generated
samples, computes the RL training loss using the prepared signals, and
updates its parameters. The updated weights are then used by the next
generation stage.
% Using the outputs from the previous stages, the loss function is computed, and the parameters of both the actor and the critic are updated accordingly. The updated actor is then used in the next iteration’s generation stage.

% Fig.~\ref{fig:rlhf-workflow} also identifies various data dependencies across the three stages. For example, stages 2 and 3 depend on the outputs of stage 1, whereas stage 1 requires the model parameters updated in stage 3 to perform generation in the subsequent iteration. Consequently, all three stages must be executed sequentially. 
% To reduce data movement and synchronization overhead between different stages, mainstream RLHF frameworks such as VeRL~\cite{sheng2025hybridflow} and RLHFuse~\cite{zhong2025optimizing} usually employ a colocation deployment strategy, where all stages share the same set of GPU servers. Placing dependent stages on the same hardware or memory domain minimizes communication latency, enables more efficient sharing of generated samples, model weights, and auxiliary outputs across stages.
Due to these dependencies, mainstream RLHF frameworks such as VeRL~\cite{sheng2025hybridflow}
and RLHFuse~\cite{ouyang2022training} adopt colocated synchronous execution, where
all stages share the same GPU pool. This deployment reduces data movement
and synchronization overhead across stages.
\subsection{Parallelism Strategies}
To improve throughput and reduce per-device memory pressure, RLHF training adopts hybrid combinations of existing parallelism strategies across its three stages~\cite{sheng2025hybridflow, mei2024realhf}. Among these strategies, Data Parallelism (DP) replicates the model across multiple devices, with each replica processing a different subset of the batch. During training, gradients are synchronized across replicas after each iteration~\cite{li2020pytorch}. 
Tensor Parallelism (TP) splits model layers across devices, allowing computations within a single layer to be executed in parallel and reducing per-device memory usage~\cite{shoeybi2019megatron}. 
% Because TP introduces substantial inter-device communication overhead, it is prevalently confined within a single server to leverage the high-bandwidth local links, such as NVLink.
Pipeline Parallelism (PP) partitions the model into sequential stages on different devices, passing micro-batches through the stages in a pipelined manner to overlap computation and communication~\cite{huang2019gpipe}. 
% It suffers from ``pipeline bubbles", which reduce overall GPU utilization.
% To improve throughput and reduce per-device memory pressure, LLMs typically employ a combination of parallelism strategies during both training and inference.
% \subsubsection{Data Parallelism}
% The full set of model parameters are replicated across multiple devices. The training dataset is partitioned into subsets, each assigned to a device for independent forward and backward computation. Before parameter updates, all devices synchronize gradients via all-reduce to ensure model consistency.
% \subsubsection{Tensor Parallelism}
% Tensors involved in computation are partitioned across multiple devices, thereby reducing memory consumption on each individual GPU. However, this approach introduces substantial inter-device communication overhead.
% \subsubsection{Pipeline Parallelism}
% The model is divided into stages based on its layer structure and stages are assigned to different devices. The input batch is further split into microbatches to enable pipelined execution. While pipeline parallelism incurs relatively low communication overhead, it suffers from unavoidable ``pipeline bubbles", which reduce overall resource utilization.

%%CL Among the three stages of RLHF, the training and preparation stages are both computation-intensive tasks, thereby employ a combination of the above three parallelism strategies (aka 3D parallelism). 
The three main stages of RLHF have distinct computational demands, necessitating different parallelism strategies~\cite{sheng2025hybridflow}. The preparation and training stages are typically compute-intensive and can benefit from hybrid parallelism, including DP, TP, and PP, depending on model size and hardware scale.
%%CL In contrast, the decoding phase of the generation stage is memory-bandwidth-bound, and GPU utilization can be improved primarily by increasing the batch size. 
In contrast, the decoding phase is primarily memory-bandwidth-bound. Its performance is optimized by maximizing batch size rather than computational throughput. Therefore, the decoding phase uses a combination of DP and TP, while favoring the largest feasible DP degree to maximize the number of concurrent decoding requests and thus improve overall throughput. PP is usually avoided during decoding because autoregressive generation produces only one token per step, so the active batch is already the main source for amortizing weight loading, KV cache access, and kernel overhead. Further micro-batching fragments the decoding workload, reduces memory-access efficiency, and may hurt overall throughput.
% Finally, as the prefill and decoding phases of generation are deployed on the same devices, they share the same parallelism strategy.
% Given the complementary strengths and weaknesses of these approaches, state-of-the-art large scale training and inference frameworks, such as Megatron and Megascale, adopt a 3D parallelism strategy that integrates data parallelism, tensor parallelism, and pipeline parallelism to balance efficiency and resource utilization.

\subsection{Problems in Synchronous RLHF Training}
% \begin{figure}[htbp]
% \centerline{\includegraphics[width=0.95\linewidth]{fig/step_time_verl.pdf}}
% \caption{End-to-end RLHF iteration latency with increasing maximum generation length.}
% \label{iteration-latency}
% \end{figure}

\begin{figure}[!t]
  \centering
  \subfloat[End-to-end iteration latency with increasing max generation length.]{%
    \includegraphics[width=0.485\columnwidth]{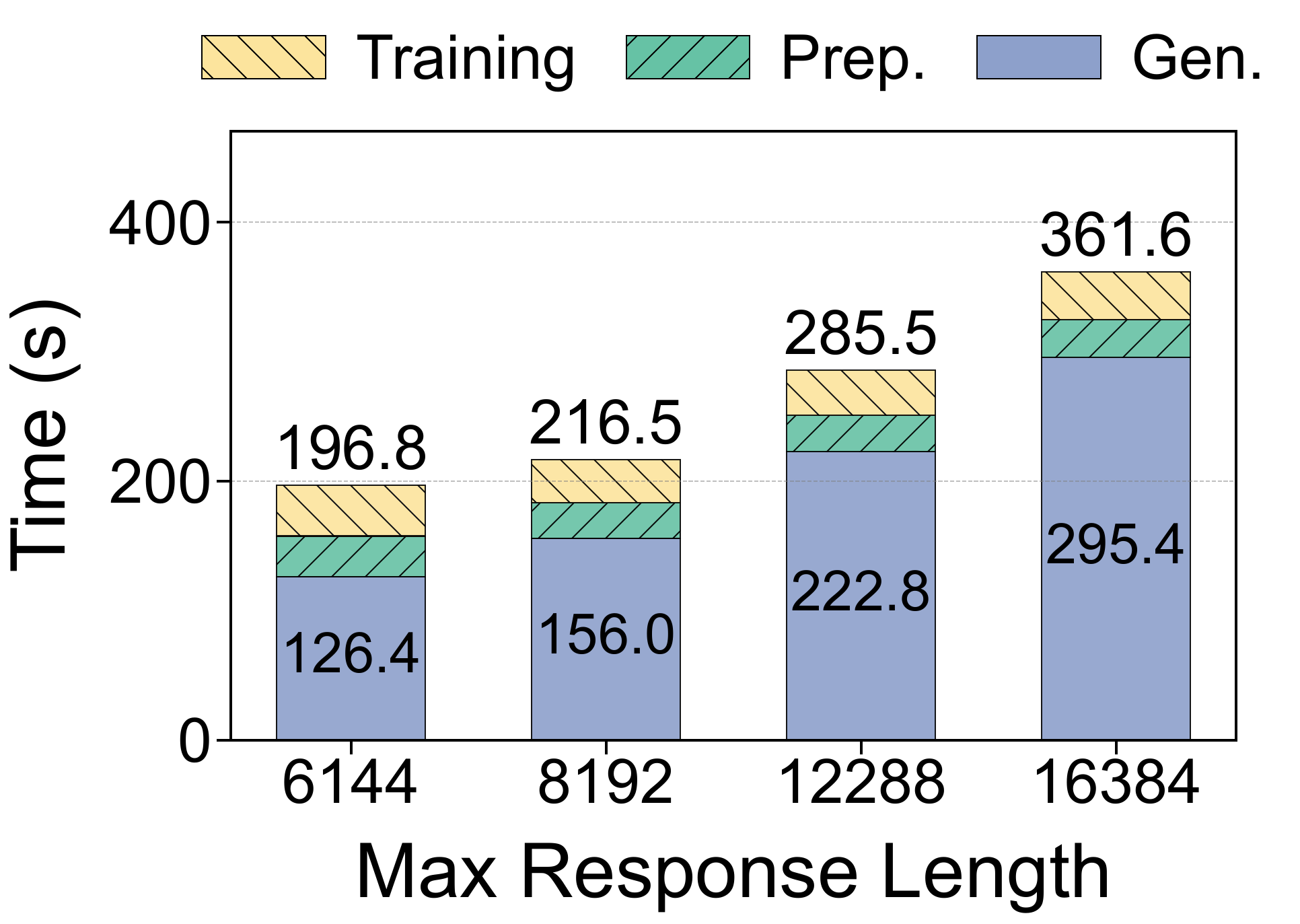}\label{fg2.1a}}%
  \hspace{1mm}% <--- 这里的 % 非常关键，确保没有多余空格
  \subfloat[CDF of response lengths generated by LLaMA3.1-8B on DeepScaleR.]{%
    \includegraphics[width=0.485\columnwidth]{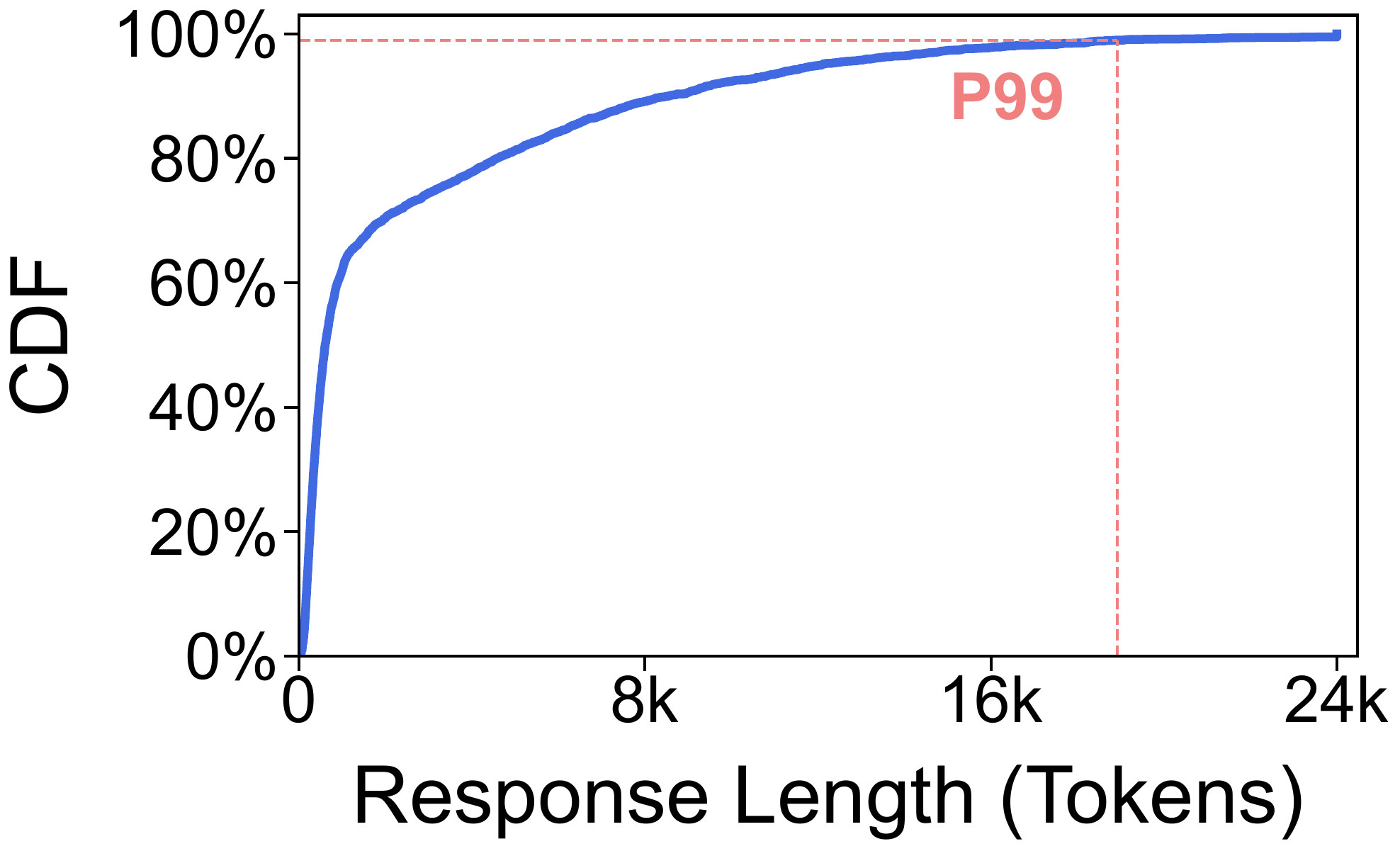}\label{fg2.1b}}
\caption{RLHF iteration breakdown and response-length distribution.}
\label{e2e&cdf}
\end{figure}

The generation stage is the primary bottleneck in synchronous RLHF training. To quantify this, we configure a tuned static parallelism strategy for each stage and follow the HybridFlow~\cite{sheng2025hybridflow} setup with a batch size of 128, using LLaMA3.1-8B on the DeepScaleR~\cite{luo2025deepscaler} dataset over eight NVIDIA A40 GPUs. As shown in Fig.~\ref{fg2.1a}, the generation stage consumes 64.22\% to 81.7\% of the total iteration time, and its share increases with the maximum response length. Across the evaluated maximum response lengths and TP/DP configurations, decoding accounts for over 90\% of the generation latency.

This decoding bottleneck mainly comes from response-length skew within each batch: a small fraction of sequences can be substantially longer than the rest~\cite{zhong2025optimizing}. 
Once shorter sequences finish, the active batch size drops quickly, but the generation stage must still wait for the remaining long sequences to complete. 
These few tail sequences therefore become the critical path of generation, leaving GPUs severely underutilized~\cite{zhong2025streamrl}. 
For example, across the maximum response lengths evaluated in Fig.~\ref{fg2.1a}, the period with only a single remaining sample accounts for 46.29\%--81.58\% of the generation latency. In the 16K setting with a batch size of 128, the average per-GPU achieved throughput drops from 13.71 TFLOPS in the aligned phase to only 0.11 TFLOPS near the end of generation.

% The key cause is the long-tail distribution of response lengths in decoding batches: a small fraction of sequences can be much longer than the rest~\cite{zhong2025optimizing}. While these long sequences are still being decoded, most other sequences have already finished, causing the effective batch size to drop sharply. As a result, the remaining few long-tail sequences become the critical path of generation and leave GPUs severely underutilized~\cite{zhong2025streamrl}. For example, across the evaluated maximum response lengths in Fig.~\ref{fg2.1a}, the tail period after only one sample remains accounts for 46.29\%--81.58\% of the generation latency. In the 16K setting with the same batch size of 128, the average per-GPU achieved throughput drops from 13.71 TFLOPS in the early decoding phase to only 0.11 TFLOPS near the end of generation.

This long-tail behavior is not unique to a specific dataset. Prior RLHF systems have also identified long-tail generation as a major source of stage-level inefficiency~\cite{zhong2025optimizing,zhong2025streamrl}. Such behavior commonly appears in workloads with substantial response-length variability, including reasoning, code generation, and instruction following. To quantify this effect, we further analyze the response length distribution of LLaMA3.1-8B on DeepScaleR. As shown in Fig.~\ref{fg2.1b}, 10.85\% of sequences exceed 8K tokens, and 2.18\% are extreme outliers exceeding 16K tokens. With a common batch size of 128, this distribution implies that each batch contains 2.8 such extreme long-tail sequences on average, which is sufficient to create a severe tail-phase bottleneck. These observations motivate \our{} to accelerate tail phase decoding in RLHF workloads with non-negligible response-length skew.

Existing systems such as StreamRL~\cite{zhong2025streamrl} mitigate long-tail bubbles by asynchronously overlapping generation with other stages. However, such approaches may introduce stale-rollout trade-offs that affect training accuracy. In contrast, \our{} accelerates the tail phase within the synchronous RLHF execution flow, without changing the RLHF algorithmic semantics.

%% file: 3.observation.tex
\section{Design Rationale}

\begin{figure}[!t]
\centerline{\includegraphics[width=0.90\linewidth]{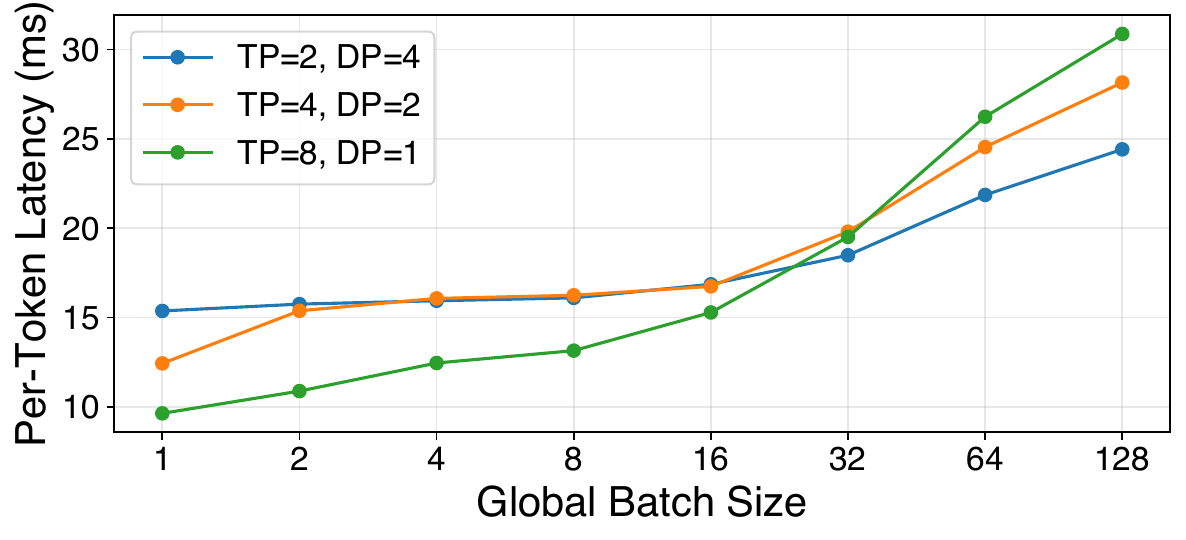}}
\caption{Impact of TP degree and global batch size on per-token decoding latency.}
\label{various-tp-latency}
\end{figure}

\subsection{Observation and Insight}
% As illustrated in Fig.~\ref{switch-example}, we break the generation stage into two distinct phases.   The first is the \textit{aligned} phase, during which most samples with comparable response lengths complete decoding nearly simultaneously. The second is the \textit{tail} phase, comprising only a few samples producing significantly longer responses, which decode slowly and result in a performance bottleneck. 

% Having shown that long-tail decoding reduces the active batch size over time, we next examine how the preferred TP/DP configuration changes with batch size. A key observation is that using a static TP/DP configuration throughout decoding is suboptimal. The aligned phase has a large active batch size, while the tail phase handles a rapidly diminishing one, and the optimal TP degree is highly sensitive to batch size.

The shrinking active batch size raises a question: should generation continue using the same TP/DP configuration in the tail phase? To answer this, we study how decoding latency changes under different TP degrees and batch sizes. Results show that the preferred TP degree is strongly batch-dependent, making a single static TP/DP configuration inefficient across the entire decoding process.

We quantify this effect using LLaMA3.1-8B on the DeepScaleR dataset. Fig.~\ref{various-tp-latency} reports the per-token decoding latency under different TP degrees and global batch sizes, revealing a clear batch-dependent trade-off. 
When the batch size is 1, increasing TP from 2 to 8 reduces the per-token latency from 15.37 ms to 9.64 ms, a 37.3\% reduction. 
In the small-batch regime, decoding is dominated by per-token data movement from HBM to compute units. A larger TP degree shards model weights and KV states across more GPUs, reducing the data each GPU must load from HBM and increasing per-token parallelism. This reduction outweighs the extra TP communication cost.
However, when the batch size increases to 128, the same TP raises the per-token latency from 24.41 ms to 30.87 ms, a 26.5\% increase, because TP communication becomes dominant. 
These results confirm that high TP suits small-batch tail decoding, whereas a low TP better serves large-batch aligned decoding.

% To understand this relationship, we studied the impact of TP degree and batch size on per-token decoding latency using the LLaMA3.1-8B model and DeepScaleR dataset. Fig.~\ref{various-tp-latency} shows a clear trade-off between TP degree and batch size. At batch size 1, increasing the TP degree from TP2 to TP8 reduces per-token latency from 15.37 ms to 9.64 ms, corresponding to a 37.3\% reduction, because the reduction in per-rank memory traffic and computation outweighs the additional TP communication cost. In contrast, at batch size 128, the same TP increase raises per-token latency from 24.41 ms to 30.87 ms, corresponding to a 26.5\% increase, indicating that TP communication becomes dominant at large batch sizes. These results show that higher TP is beneficial for small-batch tail decoding, whereas lower TP is more efficient for large-batch aligned decoding.

% These observations motivate an adaptive strategy of dynamically switching the TP degree between the aligned phase and the tail phase within a single generation stage. Specifically, we deploy a larger TP degree for the small-batch tail phase to accelerate the remaining sequences and a smaller TP degree for the large-batch aligned phase to maximize throughput.

These observations suggest an opportunity for adaptive TP reconfiguration within a single generation stage. Specifically, a smaller TP degree can be used in the large-batch aligned phase to avoid excessive communication overhead and maximize throughput, while a larger TP degree can be used in the small-batch tail phase to reduce the latency of the remaining sequences. Such reconfiguration also requires corresponding adjustments to DP to maintain a valid parallel configuration under a fixed total number of GPUs. Fig.~\ref{switch-example} provides a concrete example with four GPUs. The static baseline configuration (TP=2, DP=2) is used throughout the entire process. Our solution uses (TP=2, DP=2) for the aligned phase and dynamically switches to (TP=4, DP=1) for the tail phase. This switch involves merging the remaining samples from separate DP groups into a unified batch, thereby increasing the effective batch size for the tail sequences and improving GPU utilization.

\subsection{Challenges of TP Switching}
\label{how-to-switch}
%%CL \youhui{In practical deployment of the generation stage, TP is typically combined with DP to leverage the available GPUs. Fig.~\ref{switch-example} illustrates an example with four GPUs, where the configuration switches from $(TP2, DP2)$ to $(TP4, DP1)$. In this case, increasing the TP degree reduces the latency of long-tail responses. Concurrently, samples previously distributed across different DP groups are merged into a unified one, which increases the effective batch size and enhances GPU utilization.
However, adaptive TP switching within a generation stage must address the following two key questions:

\para{When should the switch be triggered to maximize benefit?}
% Determining the optimal switching point between the aligned and tail phases is critical. Switching too early—while the batch size remains large—can increase per-token latency, as shown in Fig.~\ref{various-tp-latency}, since a higher TP degree introduces communication overhead that outweighs its benefits under substantial batch sizes. Conversely, switching too late—when only a very small number of sequences remain—limits the potential for latency reduction. Although a higher TP degree accelerates the remaining long-tail sequences, the minimal number of such sequences offers limited benefit, which may fail to amortize the switching overhead. Thus, identifying the switching point requires a careful trade-off between the latency reduction achieved through higher parallelism and the costs associated with reconfiguration. Furthermore, selecting the appropriate TP degree at the switching point is essential to maximize inference efficiency and overall system performance.
The first challenge is to decide when a TP switch is beneficial. As decoding progresses, finished samples reduce the active batch size while unfinished samples accumulate longer contexts, causing the optimal TP/DP configuration to change over time. Therefore, PAT cannot rely on a fixed threshold; it must compare the predicted remaining latency under the current and candidate configurations, including the one-time reconfiguration cost, and switch only when the benefit can amortize this cost.
% \zw{Option: Conversely, if the switch is performed too late, some potential optimization space would be wasted.}

% To ensure that the generation stage consistently operates under the optimal parallel strategy, it is critical to carefully determine the switching point. As illustrated at figure~\ref{various-tp-latency}, once computation becomes compute-bound, a larger batch size leads to significantly higher communication overhead. Switching to a larger TP size too early may thus negate potential computational gains. In contrast, postponing the switch until only a smaller batch remains is generally more beneficial. However, switching too late reduces the benefit, as the number of remaining long-tail samples may be insufficient to amortize the switching overhead. Hence, the switching point need to strike a balance between performance and cost in order to maximize overall efficiency.

% In addition, the figure demonstrates that different TP configurations exhibit distinct optimal switching points. In practice, \our{} can gradually increase the TP size to ensure that the configuration remains optimal for the current workload. Nevertheless, multiple switches introduce additional overhead, making it necessary to carefully balance the cost of reconfiguration against the performance benefits.

\para{How to perform the switch efficiently with minimal extra overhead?}
% The second challenge is to keep the reconfiguration overhead low enough so that it does not offset the benefit of adaptive TP. Changing the TP/DP degree requires reconfiguring communication groups, reshaping model-weight partitions, and preserving the execution states of unfinished samples. These states include generated tokens and KV cache: generated tokens must be preserved, while the KV cache must either be migrated to the target GPU or reconstructed through recomputation. A naive restart-based solution would rebuild communication groups, reload model weights, and reconstruct the KV cache by re-prefilling the prompt together with already generated tokens. Such a restart introduces prohibitive latency and can easily offset the benefit of adaptive TP.
% The second challenge is to minimize reconfiguration overhead so that it does not offset the benefit of adaptive TP. Changing TP/DP requires updating communication groups, reshaping weight partitions, and preserving unfinished-sample states, where KV cache must be either migrated or recomputed. A naive restart-based solution would handle these components from scratch, introducing prohibitive latency that can easily offset the benefit of switching.
The second challenge is to keep reconfiguration overhead low enough to preserve the benefit of adaptive TP. 
Changing TP/DP requires transitioning the generation context to a new parallelism layout, including communication groups, weight partitions, and unfinished-sample states.
In particular, the KV cache of unfinished samples must also be migrated or recomputed. 
A naive restart-based solution rebuilds these components from scratch, introducing prohibitive latency that can outweigh the benefit of switching.

%% file: 4.design.tex
\begin{figure}[t!]
\centerline{\includegraphics[width=0.95\linewidth]{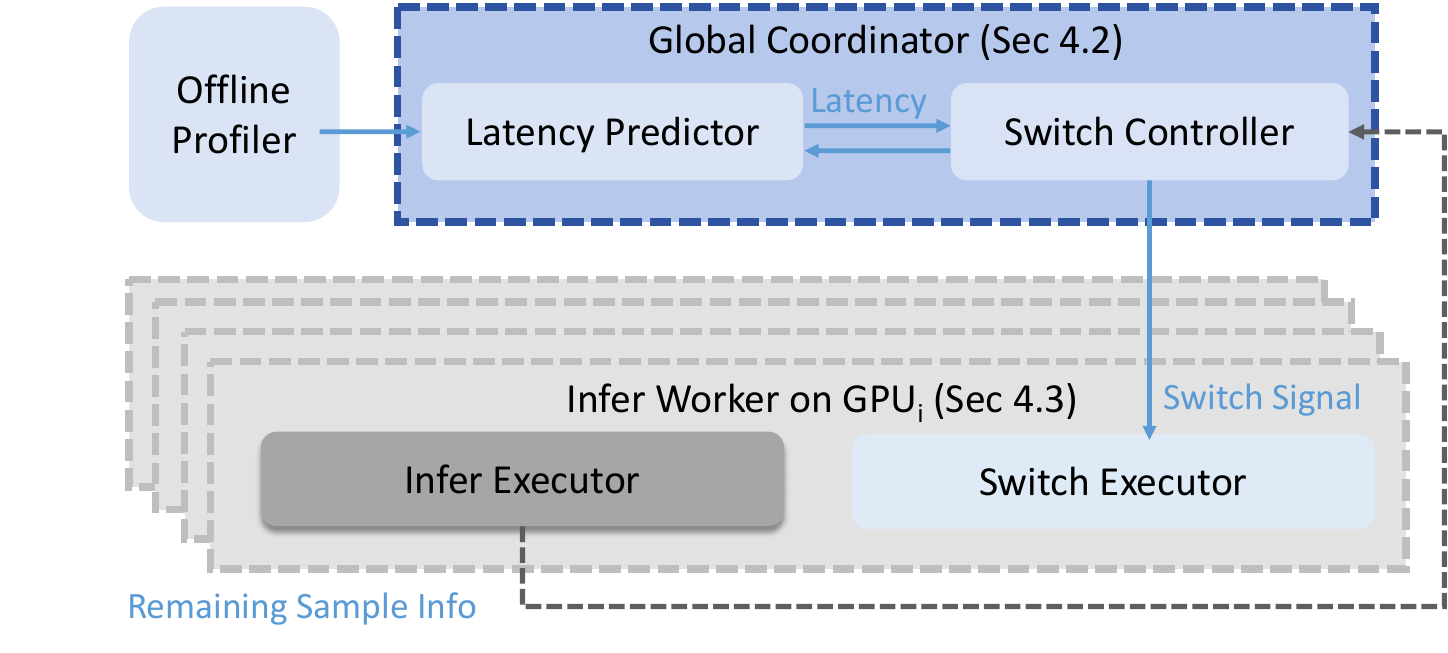}}
\caption{Architecture of \our{} with a Global Coordinator and per-GPU Inference Workers.}
\label{overview}
\end{figure}
\section{System design}
% To address the these challenges, we propose \our{}. As illustrated in figure~\ref{overview}, \our{} consists of four core components, \texttt{offline profiler}, \texttt{latency predictor}, \texttt{switch controller} and \texttt{switch executor}. \zw{Infer Executor?}

% \texttt{Offline profiler} measures the per-token decoding latency offline under different TP/DP configurations and batch sizes, given the model and hardware specifications. \texttt{Switch Controller} invokes the \texttt{Latency Predictor} based on runtime status of inference workers to estimate the remaining latency and decides whether switching is necessary. \texttt{Switch Executor} will execute reconfiguration upon receiving the switching signal by updating communication group and resharding model weights,  thereby enabling elastic TP switching.
\subsection{Overview}

To support adaptive TP/DP reconfiguration during RLHF generation, \our{} adopts a \texttt{Global Coordinator} and per-GPU worker architecture, as shown in Fig.~\ref{overview}. The \texttt{Global Coordinator} manages the global generation status and coordinates reconfiguration decisions across workers. It takes profiling data from the \texttt{Offline Profiler} and uses the \texttt{Latency Predictor} and \texttt{Switch Controller} to decide whether to switch and which target TP/DP configuration to use, as discussed in Sec.~\ref{when-to-switch}.
% and uses the \texttt{Latency Predictor} and \texttt{Switch Controller} to select the runtime TP/DP configuration, which is discussed in Sec.~\ref{when-to-switch}. 
% Following common RLHF deployments, \our{} uses intra-node TP and inter-node DP, allowing each node to reconfigure TP independently while preserving DP synchronization across nodes.
Following common RLHF deployments, \our{} uses intra-node TP and inter-node DP. 
This design allows each node to reconfigure TP independently, while DP synchronization ensures the generated samples from all nodes are collected before entering the subsequent stage.

% Each GPU runs an \texttt{Infer Worker}, which contains an \texttt{Infer Executor} and a \texttt{Switch Executor}. The \texttt{Infer Executor} performs normal token generation and reports the runtime status of unfinished samples, including active batch size and sequence lengths, to the \texttt{Global Coordinator}. Once a switch is triggered, the \texttt{Switch Executor} performs in-place reconfiguration without restarting the inference engine, including unfinished-sample state handling, weight resharding, and cached communication-group reuse. These low-overhead switching mechanisms are described in Sec.~\ref{how-to-switch}.
Each GPU runs an \texttt{Infer Worker}, which contains an \texttt{Infer Executor} and a \texttt{Switch Executor}. The \texttt{Infer Executor} performs normal token generation and reports the status of unfinished samples, such as active batch size and context lengths, to the \texttt{Global Coordinator}. When reconfiguration is triggered, the \texttt{Switch Executor} applies the new TP/DP configuration by handling unfinished-sample states, resharding model weights, and updating communication groups. The low-overhead switching mechanisms are described in Sec.~\ref{how-to-switch}.
% \youhui{
% To address these challenges, we propose \our{}, which is a high-per\-for\-mance framework for RLHF training, adaptively adjusting the TP/DP configurations during the generation stage.
% As illustrated in Fig.~\ref{overview}, 
% \texttt{Offline Profiler} measures per-token decoding latency under different TP degrees before the RLHF training. 
% Deployed on the main GPU, \texttt{Latency Predictor} performs latency estimation, and \texttt{Switch Controller} initiates TP-degree switches according to these predictions.
% Across all GPUs, \texttt{Infer Executor} and \texttt{Switch Executor} handle new token generation and switch execution, respectively.

% During inference, the \texttt{Infer Executor} generates new tokens and reports the information of remaining samples to the \texttt{Switch Controller}, which invokes the \texttt{Latency Predictor} to estimate latency and determines whether to switch the new TP/DP configuration. When triggered,  \texttt{Switch Executor} handles group redistribution and data management.
% }

\subsection{Predictor-based Switch Timing Decision}
\label{when-to-switch}
% 我们的离线分析组件会在特定<model, hardware>组合上执行不同TP、Batchsize、Sequence Length下执行一次性稀疏采样过程。
% 考虑到 TP 通常不跨节点且通常为 2 的幂次，我们针对其可能的四种配置（TP 1, 2, 4, 8）进行测量。
% 我们在参数空间内进行稀疏采样，例如在 A40 集群实验中，我们对 Batch Size $B \in [1, 256]$ 采样了 10 个代表点，
% 并对序列长度 $L \in [512, 8K]$ 采样了 5 个点。测量时，我们将目标长度作为前缀注入，连续采样 60 个 Decode Token 来计算平均延迟（这与实际生成到目标长度获取decode延迟是等价的），
% 这种采样过程独立执行 5 次以消除波动。这构成了仅由 50 种配置组成的精简 Profiling 集合。
This section first presents the overall online switching strategy, which determines both when to trigger a switch and which TP/DP configuration to select during decoding. We then describe the key components that enable this decision process, including the offline profiler and the online latency predictor.

\para{Online Switch Strategy.}
\our{} employs Algorithm~\ref{alg:switch_controller} to make cost-based switching decisions during decoding. The algorithm is invoked repeatedly within one generation stage, so PAT can perform multiple TP/DP reconfigurations as the active batch size and context lengths change. Under a fixed GPU budget, selecting a target TP degree also determines the corresponding DP degree.
% to determine both the switching point and the target TP degree, with the corresponding DP degree determined by the fixed GPU budget.

During decoding, the \texttt{Infer Executor} reports the runtime status of unfinished samples to the \texttt{Switch Controller}, including the active batch size and current sequence lengths. 
The controller first estimates the remaining decoding time under the current configuration (Line~\ref{line:4}). 
It then enumerates each candidate TP degree. 
For each candidate, it computes the merged batch size after redistributing unfinished samples across DP groups (Line~\ref{line:7}), estimates the remaining decoding time under the target configuration (Line~\ref{line:8}), and adds the one-time switching overhead (Line~\ref{line:9}). 
The remaining time is estimated using the maximum response length, which provides a conservative bound: unfinished samples may stop earlier, but they cannot exceed this length. 
The switching overhead includes generation-context transition costs such as unfinished-state handling, weight resharding, and communication-group updates, which are detailed in Sec.~\ref{how-to-switch}.

After evaluating all candidates, the controller selects the TP degree with the minimum total predicted cost, i.e., remaining decoding time plus switching overhead (Lines~\ref{line:6}--\ref{line:14}). A switch is triggered only when this cost is lower than continuing under the current configuration (Line~\ref{line:15}), ensuring that the predicted benefit on tail samples can amortize the switching overhead. Once triggered, PAT pauses decoding, merges and redistributes unfinished samples, performs TP/DP reconfiguration, and then resumes decoding under the new configuration (Lines~\ref{line:23}--\ref{line:25}). Otherwise, decoding continues under the current configuration and the controller reevaluates later.
% During decoding, the \texttt{Infer Executor} continuously reports the runtime status of unfinished requests to the \texttt{Switch Controller}, including the active batch size and the current sequence lengths of unfinished samples. 
% Based on this runtime status, the \texttt{Switch Controller} invokes the \texttt{Latency Predictor} to estimate the remaining decoding time under the current configuration (Line~\ref{line:4}). 
% For each candidate TP degree, the controller estimates the remaining decoding time after reconfiguration and adds the one-time switching overhead from the current configuration. 
% This overhead includes communication-group updates, model-weight resharding, and unfinished-state handling, such as KV-cache migration or recomputation. 
% The controller then selects the target TP degree $tp^*$ that minimizes the total predicted cost, i.e., the sum of the remaining decoding time and the switching overhead (Lines~\ref{line:6}--\ref{line:14}).

% If this minimum total predicted cost is lower than the remaining decoding time under the current configuration, the switch is considered beneficial (Line~\ref{line:15}). 
% In this case, the \texttt{Switch Controller} suspends decoding, merges unfinished samples across DP groups, sends a switching signal to the \texttt{Switch Executor}, and resumes decoding once reconfiguration is completed (Lines~\ref{line:23}--\ref{line:25}). 
% Otherwise, decoding continues under the current configuration.

\begin{algorithm}[h!]
\caption{Online Switch Mechanism}
\label{alg:switch_controller}
\begin{algorithmic}[1]
\REQUIRE Current TP $tp$, candidate TP sizes $tp\_list$, remaining batch sizes for all DP groups $rbs\_list$, max response length $L_{max}$, generated length $L_{gen}$, running requests $\mathcal{R}$
\ENSURE Minimized residual decoding time

\STATE $do\_switch \gets \textbf{false}$
\STATE $tp_{new} \gets tp$

\WHILE{decoding not finished} \label{line:3}
    \STATE $T_{cur} \gets \text{est\_rem\_time}(tp, rbs\_list, L_{max}, L_{gen})$\label{line:4}
    \STATE $T_{\text{best}} \gets \infty$
    
    \FOR{each $tp' \in tp\_list$} \label{line:6}
        \STATE $bs' \gets \text{compute\_merged\_bs}(rbs\_list, tp')$ \label{line:7}
        \STATE $T_{rem} \gets \text{est\_rem\_time}(tp', bs', L_{max}, L_{gen})$ \label{line:8}
        \STATE $T_{total} \gets T_{rem} + T_{switch}(tp, tp')$ \label{line:9}
        \IF{$T_{total} < T_{\text{best}}$}
            \STATE $T_{\text{best}} \gets T_{total}$
            \STATE $tp_{new} \gets tp'$
        \ENDIF
    \ENDFOR \label{line:14}
    
    \IF{$tp_{new} \neq tp$ \AND $T_{cur} > T_{\text{best}}$}\label{line:15}
        \STATE $do\_switch \gets \textbf{true}$
        \STATE // Exit decoding to perform TP switch
        \STATE \textbf{break}
    \ENDIF
    
    \STATE // Continue decoding
\ENDWHILE\label{line:21}

\IF{$do\_switch$}
    \STATE $\text{merge\_and\_redistribute}(\mathcal{R})$\label{line:23}
    \STATE $\text{switch\_parallel}(tp, tp_{new})$\label{line:24}
    \STATE // Resume decoding\label{line:25}
\ENDIF
\end{algorithmic}
\end{algorithm}
\para{Offline Profiler.} The \texttt{Offline Profiler} performs a one-time automated sparse-sampling process for each $\langle model, hardware \rangle$ deployment across TP degrees, batch sizes, and context lengths. Since tensor parallelism is typically confined to a single node and practical TP degrees are usually powers of two, we profile four TP configurations: $TP \in \{1, 2, 4, 8\}$.

To balance profiling efficiency and prediction accuracy, we sample only the feasible workload region. In our A40 experiments, we profile 10 representative batch sizes $B \in [1, 256]$ and context lengths $L \in [8, 128K]$. To avoid unrealistic long-context/high-batch combinations, we retain a profiling point $(B,L)$ only if $B \cdot L \le T_{\text{cap}}$, where $T_{\text{cap}}$ is the deployment-level active-token budget determined by the runtime KV cache capacity. For LLaMA3.1-8B on eight A40 GPUs, we set $T_{\text{cap}}=2^{20}$ tokens, corresponding to a conservative budget of 128K tokens per GPU. We further increase the profiling density for short contexts, e.g., $L < 512$, where decoding latency shows more pronounced nonlinear behavior.

For each profiled batch size, we measure decoding latency using dummy batches to obtain controlled profiling points with fixed shapes. Real dataset samples have irregular and uneven length distributions, making it difficult to cover the desired $(B,L)$ grid systematically. Therefore, for a target batch size $B$ and context length $L$, we randomly generate token IDs of length $L$ for each sequence and inject them as prefixes. During prefix construction, we record the corresponding prefill latency, which is later used to estimate the recomputation cost in Sec.~\ref{kv-cache-handling}. We then run the following 60 decoding steps and use their average latency as the per-token decoding latency at that context length.

This profiling avoids generating tokens from scratch up to each target length, reducing each measurement to only one prefix construction followed by a short decoding window. As a result, the total profiling time is reduced to approximately 10 minutes.

% 我们观察到两个关键的物理特性，如图 \ref{fig:latency_characterization} 所示：
% (i) 延迟与总 Token 数 $T = B \times L$ 的线性度：对于固定的 TP 大小和 $B$，延迟 $\ell$ 与 $T$ 呈近乎完美的线性关系。
% 这是由于每增加一个 Token 带来的计算开销在特定硬件密度下是恒定的。
% (ii) 延迟对 Batch Size $B$ 的非线性增长：相比之下，延迟随 $B$ 的增长虽然单调，但呈现出明显的非线性趋势，
% 这是由于现代推理后端（如 Triton 或 Flash-Infer）对特定 2 的幂次 Batch 规模具有更强的 Kernel 亲和性。
%
% 基于此，我们为每个 profiled $(TP, B)$ 组合独立拟合一个关于 $T$ 的线性模型 $\ell_{tp, B}(T) = \beta_{tp, B} \cdot T + \alpha_{tp, B}$。
% 在运行时，针对任意负载 $(B', T')$，预测器执行两级预测：
% 1. 识别 $B'$ 的相邻已 Profile 边界 $B_{low}$ 和 $B_{high}$。
% 2. 分别计算两个边界模型在 $T'$ 下的基准延迟 $\ell_{low}(T')$ 和 $\ell_{high}(T')$。
% 3. 在 BS 维度上进行线性插值得到最终结果。

% 延迟预测可扩展性探究。这种离线profiling+在线预测的设计能够很好地适应解码过程中的动态特性。尽管一个batch中的样本在运行时会随时结束导致 BS 发生变化，但是由于我们会在线汇报当前剩余样本数，
% 并且我们的离线 Profile 通过覆盖多种 BS 组合，并利用Latency Predictor对缺失bs数据进行精准拟合，因此能够很好得支持动态预测。
% 此外，Profiling 结果具有很强的跨数据集迁移能力。因为不同 LLM 数据集的区别主要在于输入输出样本的长度分布，
% 这些差异都可以量化为总 Token 数（$B \times L$）的差异，进而影响decode过程中的访存延迟，因此， 通过对这些不同token总量下的延迟数据进行拟合，我们可以预测出不同负载下的decode延迟。

\begin{figure}[t!]
  \centering
  \begin{minipage}[b]{0.48\linewidth}
    \centering
    \includegraphics[width=\linewidth]{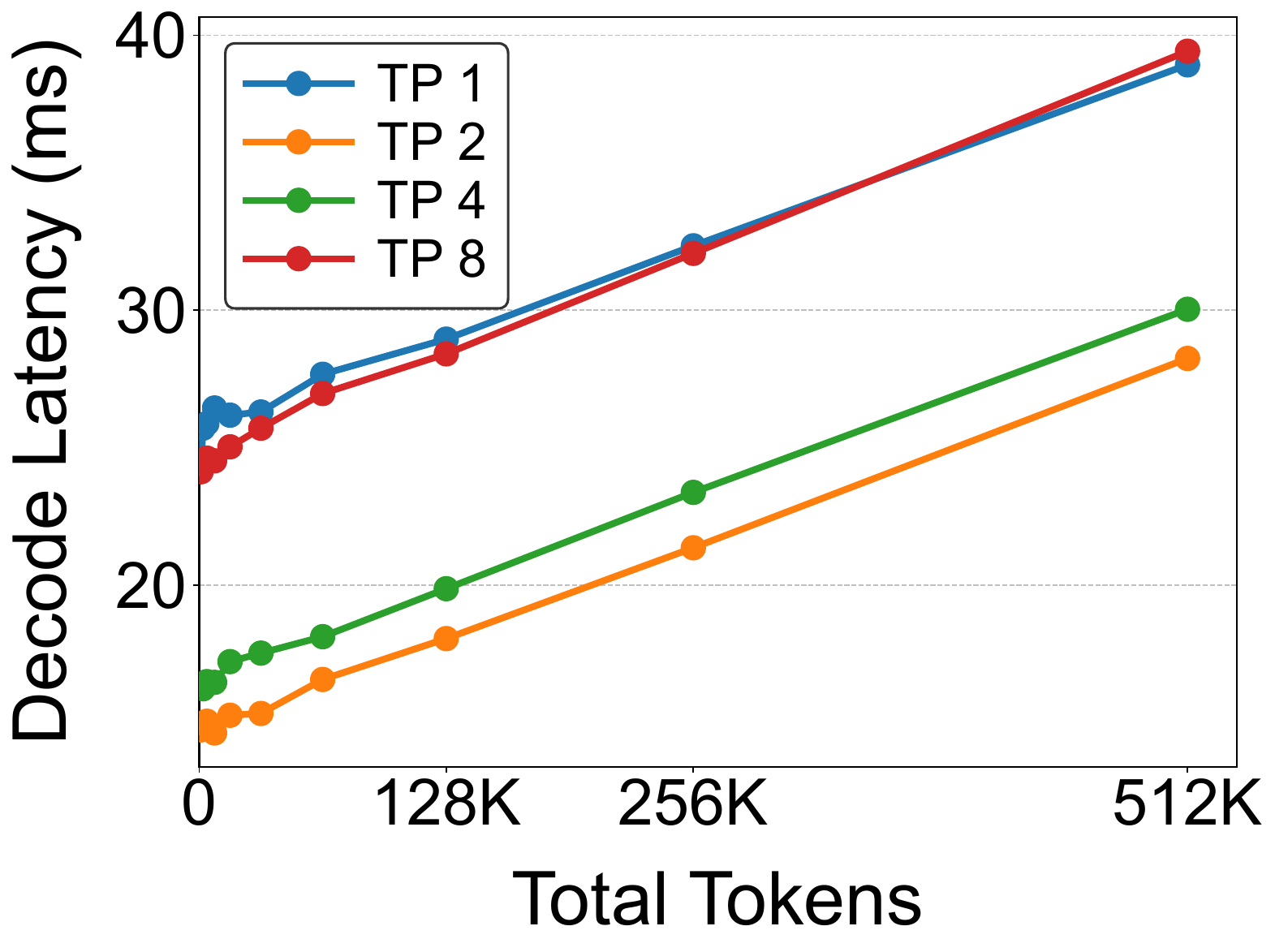}
    \centerline{\footnotesize (a) Latency vs. Total Tokens}
    \label{fig:latency_token}
  \end{minipage}
  \hfill
  \begin{minipage}[b]{0.48\linewidth}
    \centering
    \includegraphics[width=\linewidth]{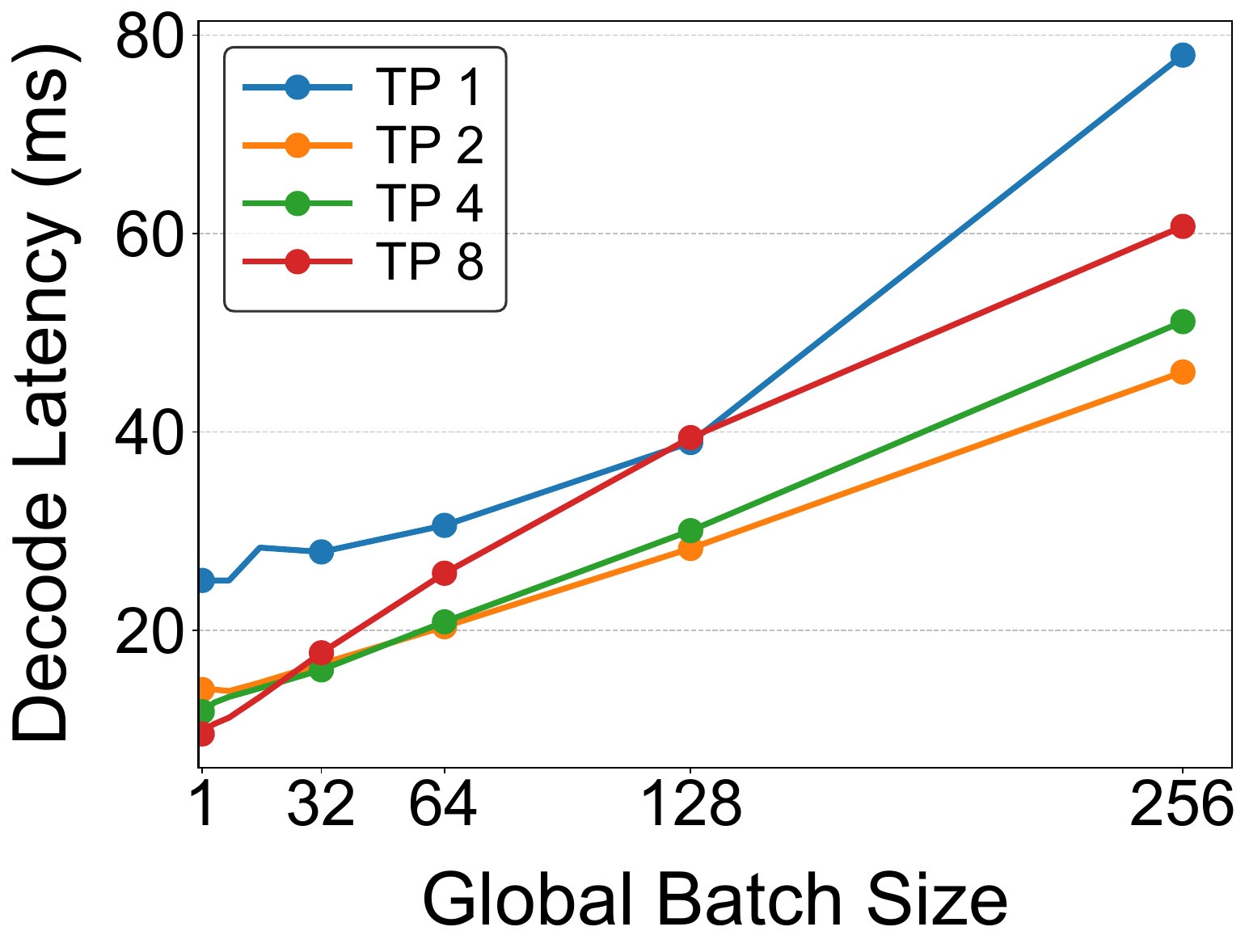}
    \centerline{\footnotesize (b) Latency vs. Batch Size}
    \label{fig:latency_gbs}
  \end{minipage}
  \caption{Characterization of decoding latency on an H100 GPU cluster under varying total tokens and batch sizes. (a) shows the near-linear scaling with respect to the aggregate token count $T=B \times L$ for a fixed batch size $B=128$. (b) shows shape-dependent nonlinear variations as the batch size $B$ changes for a fixed context length $L=4096$.}
  \label{fig:latency_characterization}
\end{figure}

\para{Online Latency Predictor.}
This module estimates two costs for the \texttt{Switch Controller}: the per-step decoding latency for a runtime workload and the one-time reconfiguration cost between TP/DP configurations.
% The \texttt{Online Latency Predictor} provides two cost estimates used by the \texttt{Switch Controller}: the per-step decoding latency under a given runtime workload and the one-time reconfiguration cost between TP/DP configurations.

As shown in Fig.~\ref{fig:latency_characterization}, decoding latency is mainly determined by the active batch size $B$ and the aggregate context token count $T=\sum_{i=1}^{B}(L_i^{prompt}+L_i^{gen})$. 
Although latency generally increases with $T$, it also varies nonlinearly across batch sizes due to kernel tiling, hardware alignment, and backend-specific optimizations. 
To capture these effects efficiently, PAT builds a piecewise linear predictor from sparse profiling data. 
At runtime, for a workload $(B,T)$ under a candidate TP degree, the predictor queries the fitted latency curves of the two nearest profiled batch sizes and interpolates along the batch dimension to obtain the per-step latency $\ell(B,T;\mathrm{TP})$.

The predictor also estimates the one-time reconfiguration cost. 
This cost includes data-movement overheads, such as model-weight resharding and unfinished-sample state handling, and control overheads, such as CUDA Graph recapturing, memory deallocation, and garbage collection. 
For unfinished-sample states, PAT uses the lower estimated cost between KV cache migration and recomputation, as detailed in Sec.~\ref{how-to-switch}. These estimates allow PAT to trigger TP/DP reconfiguration only when the expected tail-decoding benefit can amortize the transition cost.

Overall, the predictor is tied to the deployment setting rather than a specific dataset. 
Its inputs are the model, hardware/backend, TP/DP configuration, active batch size, and aggregate context tokens. 
Different datasets mainly affect the runtime values of $B$ and $T$, so the same profiled predictor can be reused across datasets under the same deployment.

\subsection{Low-overhead Parallelism Switching}
\label{how-to-switch}
% To minimize the overhead of reconfiguration, \our{} avoids process restarts and instead relies on the \texttt{switch executor} to directly update TP/DP configurations within existing worker processes, thereby  reducing switching costs. To further ensure the correct management of communication, storage, and data resources (as discussed in Section~\ref{how-to-switch}) and to reduce reconfiguration overhead, \our{} introduces an efficient switching mechanism:
% \youhui{
To mitigate the overhead of TP switching, \our{} circumvents restarting the inference engine by employing a set of reuse mechanisms with optimization trade-offs.
% }

\para{State handling for unfinished samples.}
\label{kv-cache-handling}
During TP reconfiguration, \our{} restores unfinished samples using either KV cache migration or recomputation. Migration transfers valid KV states to the target TP/DP configuration, avoiding extra prefill computation but introducing communication overhead. Recomputation transfers only token sequences and rebuilds KV states through an additional prefill pass. The \texttt{Latency Predictor} estimates both costs for each candidate configuration and uses $\min(T_{\mathrm{move}}, T_{\mathrm{recomp}})$ as the predicted state-handling cost.

\begin{figure}[t!]
  \centering
  \includegraphics[width=\linewidth]{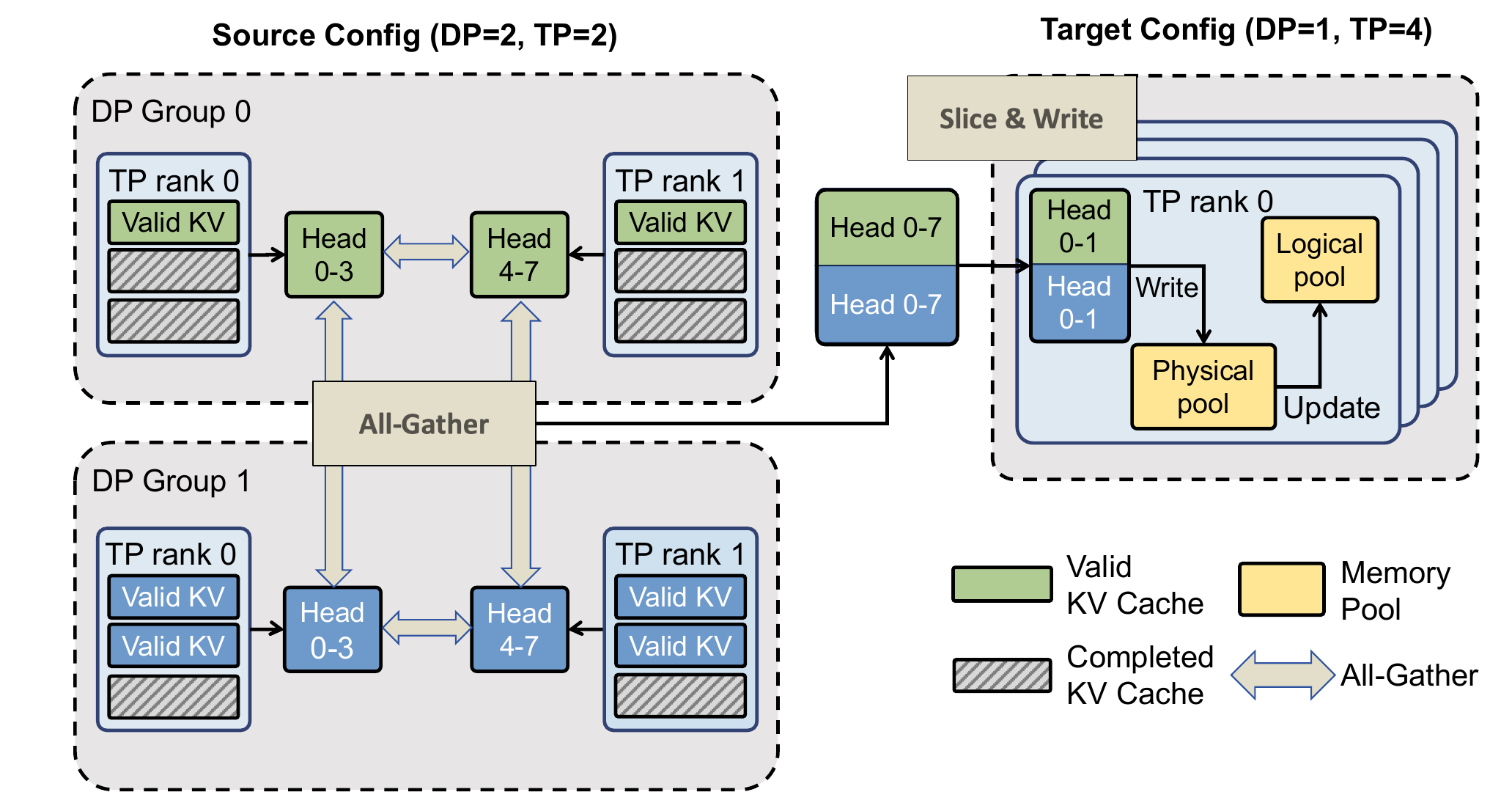}
  \caption{Workflow of KV cache migration.}
  \label{fig:kv_cache_migration}
\end{figure}

For KV cache migration, PAT estimates $T_{\mathrm{move}}$ from the amount of valid KV data on the communication critical path. As shown in Fig.~\ref{fig:kv_cache_migration}, PAT redistributes KV states using a layer-wise All-Gather and Slice procedure. GPUs first gather the valid KV shards of unfinished samples within the target TP group, and each target TP rank then slices out its required head partition. This is necessary because source KV shards are organized according to the old DP/TP layout; after DP groups are merged and TP degree is changed, a target rank may not hold the KV heads it needs locally. All-Gather and Slice reconstructs the logical KV states under the new TP group while preserving the engine’s canonical rank-to-shard mapping.

% For migration, \our{} redistributes KV states through a layer-wise All-Gather and Slice procedure, as shown in Fig.~\ref{fig:kv_cache_migration}. For each layer, GPUs gather valid KV shards of unfinished samples, and each target TP rank slices out its required head partition. We use All-Gather and Slice instead of direct local slicing because local shards are tied to the source DP/TP layout. After DP groups are merged and TP is increased, different physical ranks may hold duplicated source TP shards, whereas the target TP group expects ordered and non-overlapping head partitions. Thus, the shard required by a target TP rank may not be obtainable from its local shard alone. All-Gather and Slice avoids modifying the engine's rank-to-shard mapping: it first reconstructs the logical KV states within the new TP group and then lets each target rank extract its canonical head partition, preserving the rank semantics of engines such as SGLang~\cite{zheng2024sglang}.

% We estimate migration time from the per-rank send volume on the communication critical path. 
Let $B$ be the number of unfinished samples, $L_i$ the context length of the $i$-th unfinished sample, $M$ the number of Transformer layers, $H$ the hidden dimension, $s$ the bytes per element, and $\mathrm{TP}_{\mathrm{src}}$ the source TP degree. Since KV states are sharded across source TP ranks, the per-rank KV send volume is approximated as

\begin{equation}
S_{\mathrm{kv}}^{\mathrm{rank}} =
\sum_{i=1}^{B}
2 \cdot M \cdot L_i \cdot \frac{H}{\mathrm{TP}_{\mathrm{src}}} \cdot s ,
\end{equation}
where the factor 2 accounts for key and value tensors. Given the measured one-way per-rank bandwidth $B_{\mathrm{net}}^{\mathrm{uni}}$ under the same migration pattern, PAT estimates
\begin{equation}
T_{\mathrm{move}} =
\frac{S_{\mathrm{kv}}^{\mathrm{rank}}}{B_{\mathrm{net}}^{\mathrm{uni}}}.
\end{equation}
Here, $B_{\mathrm{net}}^{\mathrm{uni}}$ denotes one-way per-rank bandwidth rather than bidirectional aggregate PCIe/NVLink bandwidth.

For recomputation, \our{} reruns batched prefill for all unfinished samples under the target TP/DP configuration. Because unfinished samples may have different context lengths while the \texttt{Offline Profiler} records the latency of fixed-length batches, we approximate the variable-length batch with an equivalent fixed-length batch using the root-mean-square (RMS) context length. Let $\tilde{L}_i$ be the full context length of the $i$-th unfinished sample:
\begin{equation}
L_{\mathrm{RMS}} =
\sqrt{
\frac{1}{B}
\sum_{i=1}^{B}
\tilde{L}_i^2
}.
\end{equation}
This preserves the dominant quadratic attention cost in prefill, i.e., $B L_{\mathrm{RMS}}^2 = \sum_{i=1}^{B}\tilde{L}_i^2$. The recomputation time is estimated as
\begin{equation}
T_{\mathrm{recomp}} =
T_{\mathrm{prefill}}^{\mathrm{prof}}
\left(
B, L_{\mathrm{RMS}}; \mathrm{TP}_{\mathrm{target}}
\right),
\end{equation}
where $T_{\mathrm{prefill}}^{\mathrm{prof}}(B,L;\mathrm{TP})$ denotes the profiled prefill latency for batch size $B$ and context length $L$ under TP configuration $\mathrm{TP}$.
\begin{figure}[t!]
  \centering
  \includegraphics[width=1\linewidth]{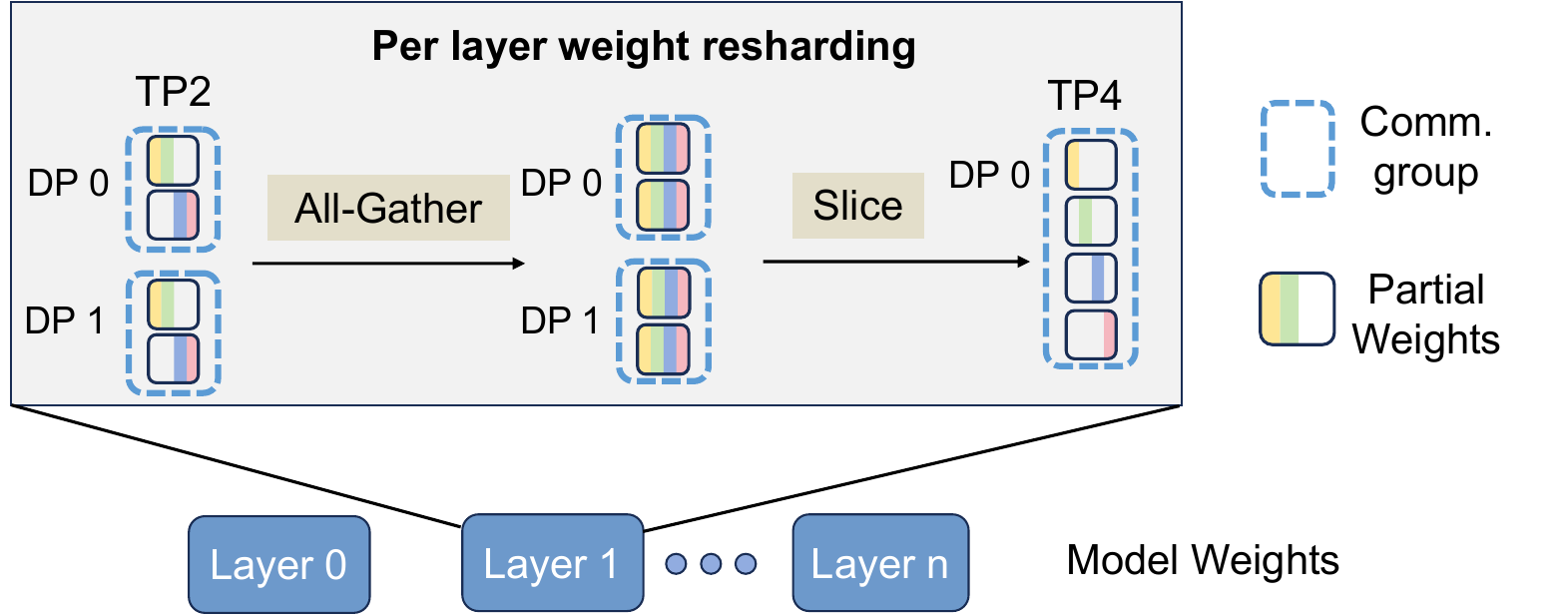}
  \caption{Reload-free layer-wise weight resharding.}
  \label{fig:weight_slicing_mechanism}
\end{figure}

\para{Efficient weight resharding.} Varying the TP degree leads to different sharding granularities of model weights across GPUs. 
A na\"ive approach to handle this transition is to reload weights from the training engine into the inference engine. 
However, format mismatches introduce substantial overhead from loading, conversion, and resharding, which is further exacerbated by CPU-to-GPU data transfers under parameter offloading. 
We observe that during the same generation stage, the weight contents remain unchanged, and only their partitioning shapes change with the TP degree.

Based on this observation, \our{} employs a reload-free, layer-wise resharding strategy, as illustrated in Fig.~\ref{fig:weight_slicing_mechanism}. 
To transition between TP/DP configurations efficiently, we implement a communication-based redistribution mechanism using a sequence of layer-wise All-Gather and Slice operations. 
Instead of a monolithic transfer, we process weights layer-by-layer to strictly bound the peak memory usage, ensuring that the resharding process does not trigger out-of-memory (OOM) errors even under high memory pressure.

% One might argue that weight resharding could be avoided by utilizing a subset of the weights based on the higher DP and lower TP degree. 
% However, the local slicing approach requires shuffling the TP rank order to change the role of each physical GPU. 
% As illustrated in Fig.~\ref{fig:weight_slicing_mechanism}, if a TP4 configuration attempts to reuse the redundant weights from a previous TP2 setup, the TP rank sequence would need to be reordered (e.g., from $[0, 1, 0, 1]$ to $[0, 2, 1, 3]$) to maintain local shard availability. 
% Redefining these rank mapping rules within mature inference engines like SGLang~\cite{zheng2024sglang} would be highly intrusive and could introduce subtle logical errors in the established communication protocols. 
% In contrast, \our{} employs explicit weight resharding to avoid rank shuffling. Moreover, it aligns with our communication group reuse design, allowing us to seamlessly leverage cached TP groups from previous iterations.
% One alternative is to reuse local shards and skip explicit resharding where the required target shards already reside on the corresponding GPUs. However, such local reuse only works for specific TP switching and requires changing the canonical mapping between physical GPUs, TP ranks, and weight partitions. This is intrusive to mature inference engines such as SGLang~\cite{zheng2024sglang}, where the weight layout, communication groups, and CUDA Graph execution all rely on a consistent TP rank order. 
% \our{} therefore performs explicit weight resharding while preserving the original TP rank mapping.
Although some local shards may already match the target layout, relying on such reuse would require changing the engine’s canonical rank-to-shard mapping. Since weight layout, TP collectives, and CUDA Graphs all assume a fixed order between TP ranks and tensor partitions, PAT explicitly reshards weights to preserve SGLang’s rank semantics.

\para{Reuse of communication groups.} 
Updating the TP degree requires reconfiguring communication groups to ensure that distributed model weight shards can correctly execute collective operations. 
To avoid the overhead of repetitive initialization, \our{} implements a \texttt{Cache Manager} that maintains a \texttt{Communication Group Pool} in each worker process.

\begin{figure}[t!]
  \centering
  \includegraphics[width=0.9\linewidth]{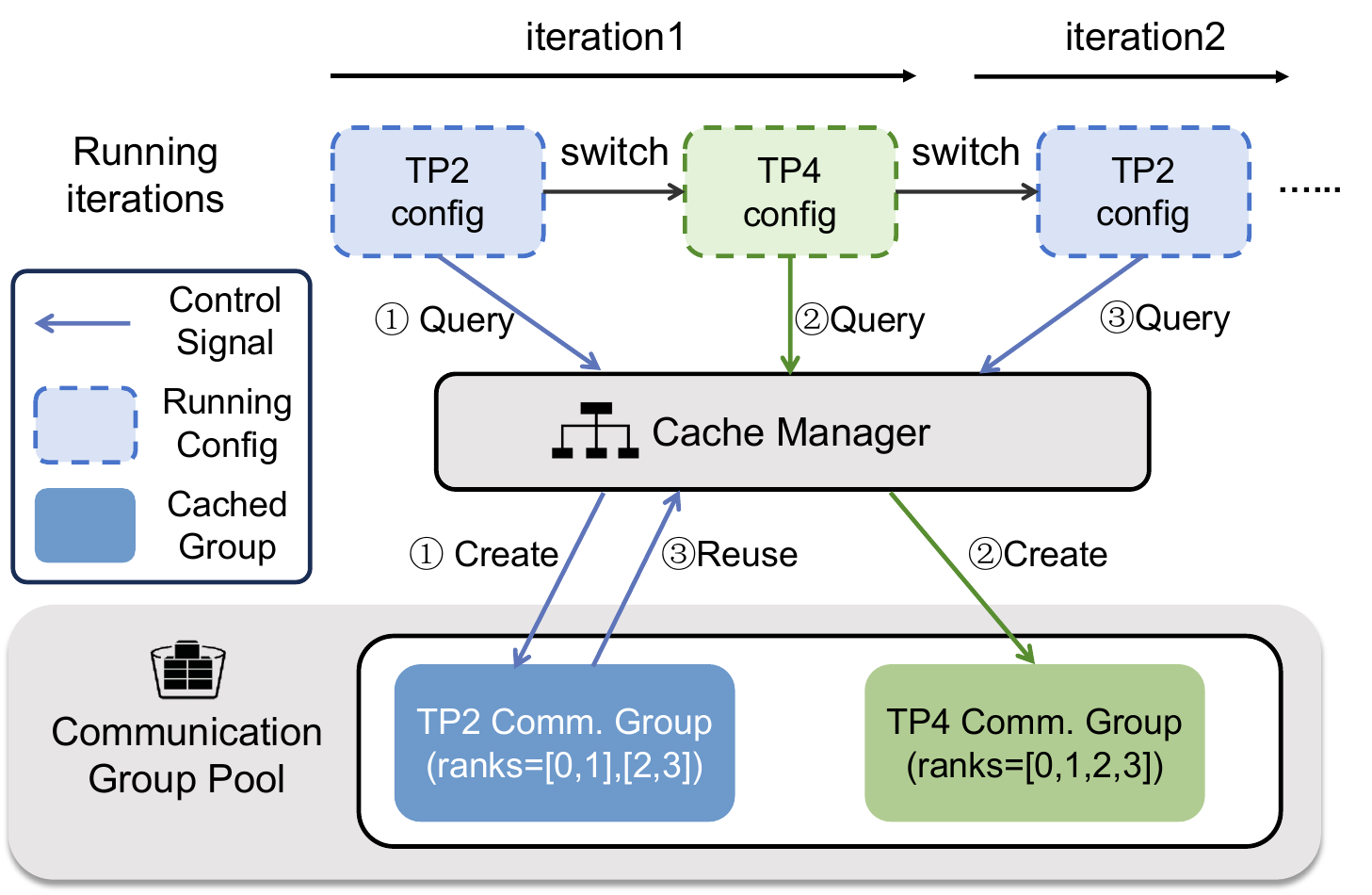}
  \caption{Lifecycle of communication group reuse. The \texttt{Cache Manager} avoids
redundant initialization by caching instantiated process groups across iterations.}
  \label{fig:comm_group_reuse}
\end{figure}

As illustrated in Fig.~\ref{fig:comm_group_reuse}, the lifecycle of a communication group involves a \textit{lazy initialization and persistent caching} strategy. 
% During the first iteration of RLHF training, if a switch from TP2 to TP4 is triggered, the \texttt{Cache Manager} encounters a cache miss for these configurations and initiates the one-time creation of the corresponding communication groups. 
When a TP/DP configuration is requested for the first time, the \texttt{Cache Manager} creates the corresponding process groups and stores their handles in the pool.
In subsequent iterations, the required communication groups are directly retrieved from the pool via cache hits. 
Consequently, over the full lifecycle of RLHF training, only the first appearance of a specific TP/DP configuration incurs group initialization cost.
Given that tensor parallelism is typically restricted to a single node, the number of candidate configurations is strictly limited (e.g., $TP \in \{1, 2, 4, 8\}$). 
Empirical measurements on our A40 single-node experiments show that the memory footprint for maintaining these idle communication handles is negligible (contributing only an additional 2\% of GPU memory usage).

\para{Correctness-preserving reconfiguration.}
The above reconfiguration preserves the original RLHF semantics. 
PAT does not change the rollout policy, reward computation, loss function, or model update rule; it only changes tensor placement across GPUs. 
In the migration path, KV cache handling and weight resharding use tensor copy, All-Gather, Slice, and metadata updates, which only reorder existing shards without arithmetic reduction, accumulation, quantization, or precision conversion. Thus, they introduce no additional floating-point error. 
In the recomputation path, PAT rebuilds KV states by running the same prefill computation with the same token sequence and model weights, matching the numerical behavior of the original backend. 
Overall, PAT preserves synchronous RLHF semantics without introducing extra numerical approximation beyond standard GPU execution.
% The above reconfiguration preserves the original RLHF semantics. \our{} does not change the rollout policy, reward computation, loss function, or model update rule. Instead, TP/DP reconfiguration only changes the physical placement of existing tensors across GPUs. In particular, KV cache migration and weight resharding are implemented through tensor copy, All-Gather, and Slice operations, without arithmetic reduction, accumulation, quantization, or precision conversion. Therefore, \our{} does not introduce additional numerical approximation beyond the original backend, and preserves the synchronization between generated samples and the current model weights.

\subsection{Implementation Details}

\our{} is built on top of VeRL~\cite{sheng2025hybridflow} and SGLang~\cite{zheng2024sglang}. 
VeRL orchestrates the RLHF workflow, while SGLang serves as the generation engine where runtime TP/DP reconfiguration is performed.

\para{Adaptation to RL framework.}
On the VeRL side, we add 895 lines of code to adapt the generation orchestration. The original VeRL generation stage implements DP by launching multiple independent inference engines, which works for static execution but prevents coordinated TP/DP reconfiguration due to isolated runtime states and process management. 
\our{} therefore replaces the multi-engine generation mode with a unified single-engine mode, allowing all inference workers to be managed under one SGLang engine instance. 
This change is an enabling mechanism for coordinated reconfiguration rather than an independent optimization: the generation kernels, batching policy, and static TP/DP execution remain unchanged when switching is disabled.

\para{Adaptation to inference engine.}
On the SGLang side, we add about 2,650 lines of code to support online TP/DP switching, including runtime reconfiguration, CUDA Graph management, and reuse of configuration-independent inference modules.

A practical challenge is that inference engines commonly capture CUDA Graphs for different batch sizes to accelerate decoding. When \our{} switches TP/DP configurations, the shapes of input tensors and weight shards change, so CUDA Graphs captured under the original configuration become invalid and cannot be directly reused. Recapturing graphs for all possible batch sizes would introduce noticeable overhead from warm-up, graph capture, and runtime variable initialization. To reduce this overhead, \our{} only captures CUDA Graphs for small batch sizes that may appear after switching in the tail phase, while avoiding unnecessary graph construction for large aligned-phase batches.

In addition, we avoid reinitializing runtime modules whose states are independent of the TP/DP configuration, such as the tokenizer and grammar backend. These modules are reused across reconfigurations, reducing switching overhead without affecting the generation kernels, batching policy, or decoding semantics.

% We implement PAT on top of VeRL and SGLang with 895 lines of code for the latency predictor and switch controller and about 2650 lines for the online switching mechanism. 
% Because the original VeRL generation stage launches multiple isolated SGLang engines, we replace it with a single-engine mode to enable shared runtime state and adaptive TP reconfiguration. 
% Since switching changes input and weight shapes, existing CUDA Graphs cannot be reused; PAT therefore recaptures graphs only for small post-switch batches, which is sufficient for long-tail decoding.

%% file: 5.evaluation.tex
\begin{figure*}[!t]
  \centering
  \makebox[\textwidth][c]{%
    \subfloat[Throughput of LLaMA3.1-8B on 8$\times$A40]
    {\includegraphics[width=0.62\columnwidth]{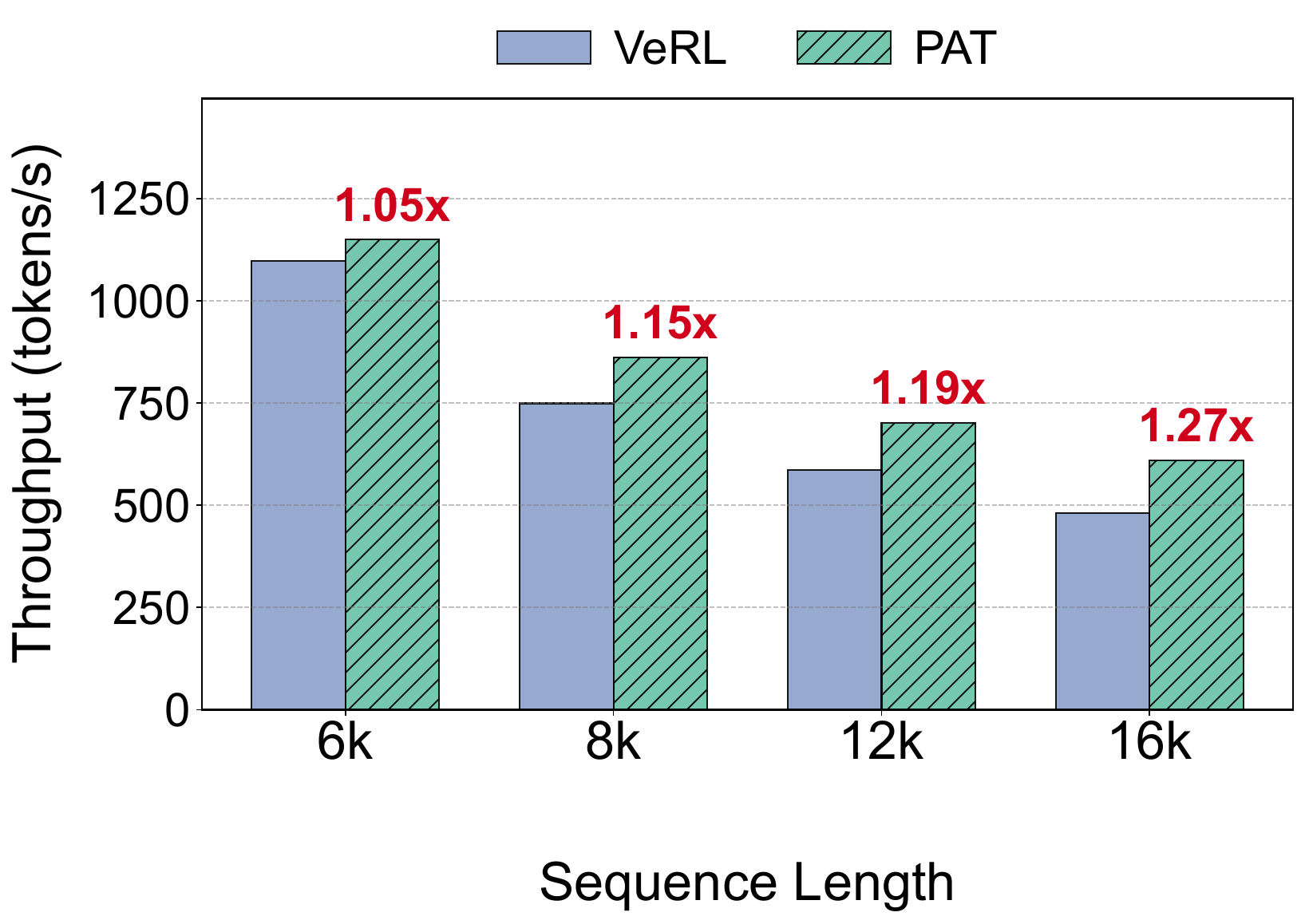}\label{fig:a40_8b_single}}
    \hspace{0.018\textwidth}
    \subfloat[Throughput of LLaMA3.1-8B on 16$\times$A40]
    {\includegraphics[width=0.62\columnwidth]{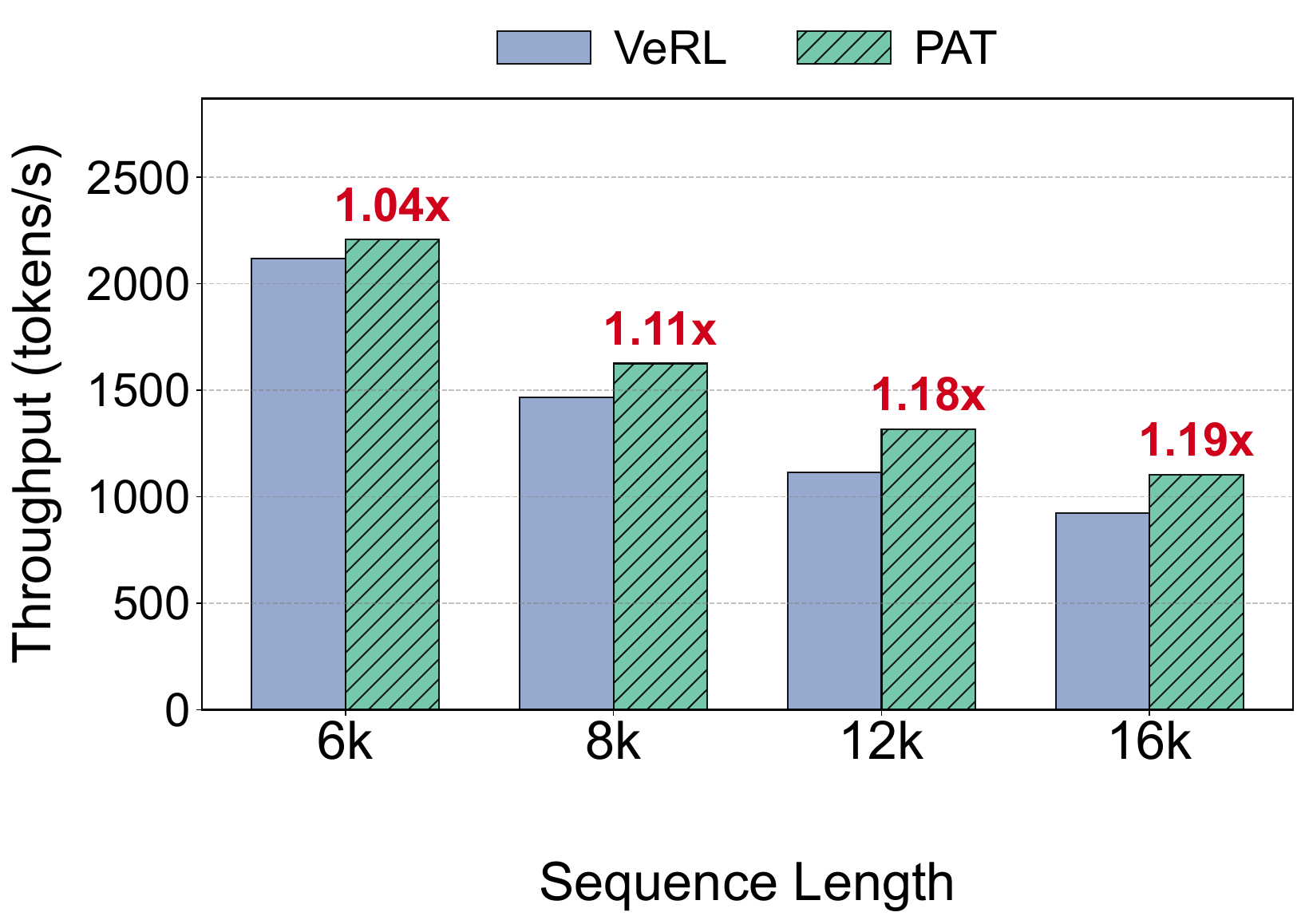}\label{fig:a40_8b_dual}}
    \hspace{0.018\textwidth}
    \subfloat[Throughput of Qwen3-14B on 16$\times$A40]
    {\includegraphics[width=0.62\columnwidth]{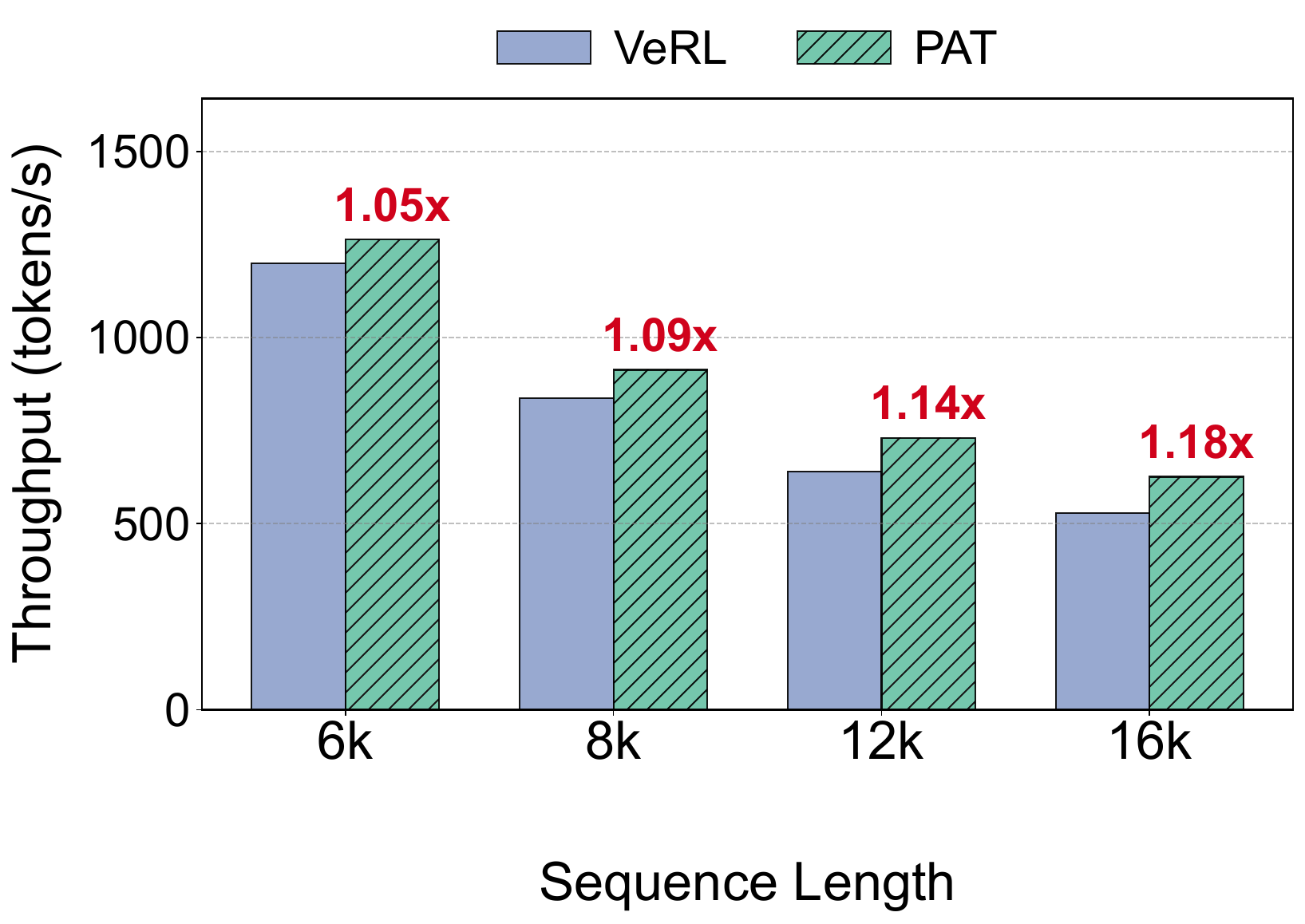}\label{fig:a40_14b_dual}}
  }
  \caption{End-to-end throughput comparison between VeRL and \our{} on the A40 cluster. (a) and (b) compare the end-to-end throughput under increasing sequence lengths for LLaMA3.1-8B on single-node and dual-node, respectively. (c) compares the end-to-end throughput under increasing sequence lengths for Qwen3-14B on dual-node.}
  \label{fig:a40_comparison}
  \vspace{-0.8em}
\end{figure*}

\begin{figure*}[!t]
  \centering
  \makebox[\textwidth][c]{%
    \subfloat[End-to-end throughput of LLaMA3.1-8B]
    {\includegraphics[width=0.62\columnwidth]{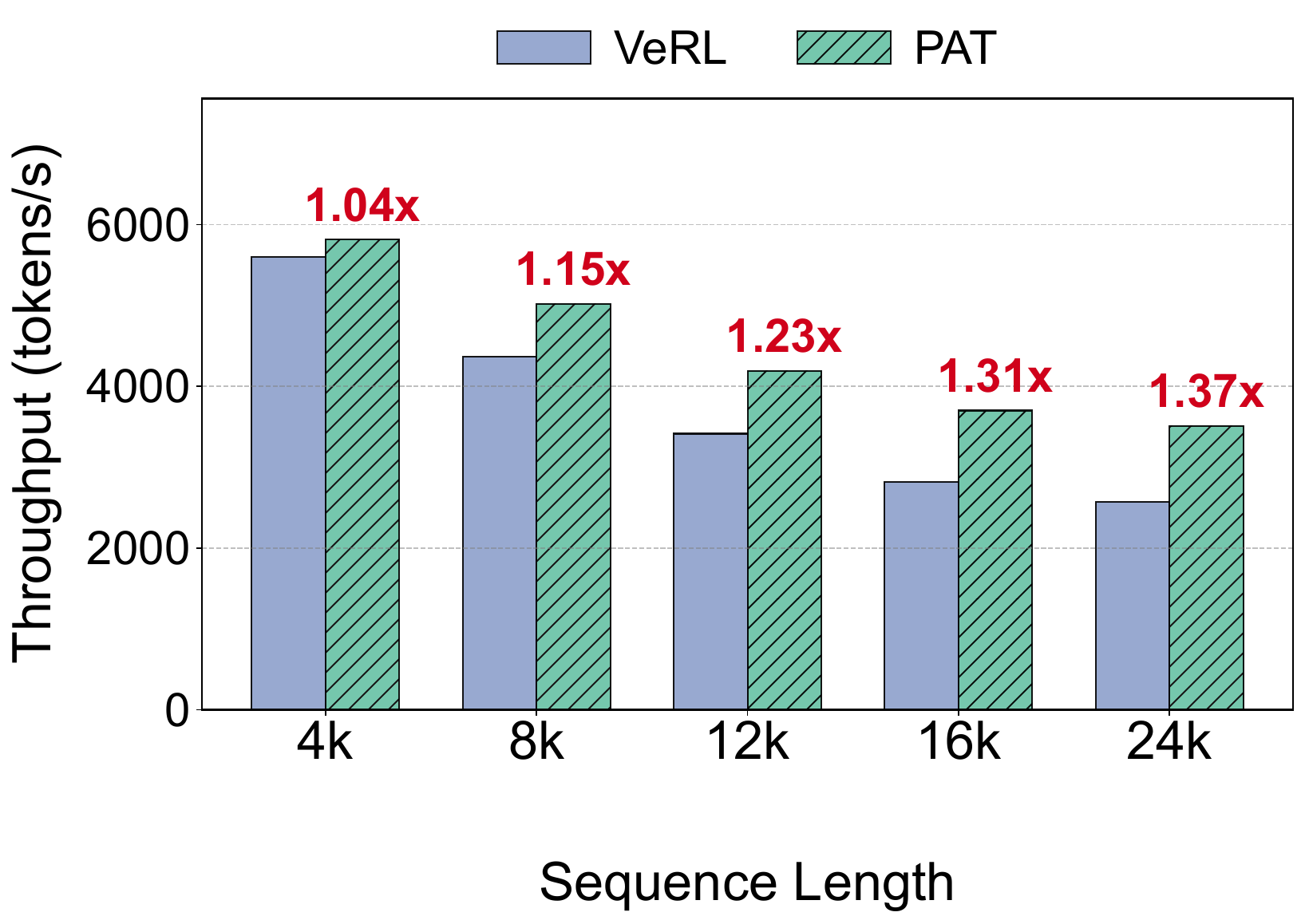}\label{fig:h100_8b_e2e}}
    \hspace{0.018\textwidth}
    \subfloat[End-to-end throughput of Qwen3-14B]
    {\includegraphics[width=0.62\columnwidth]{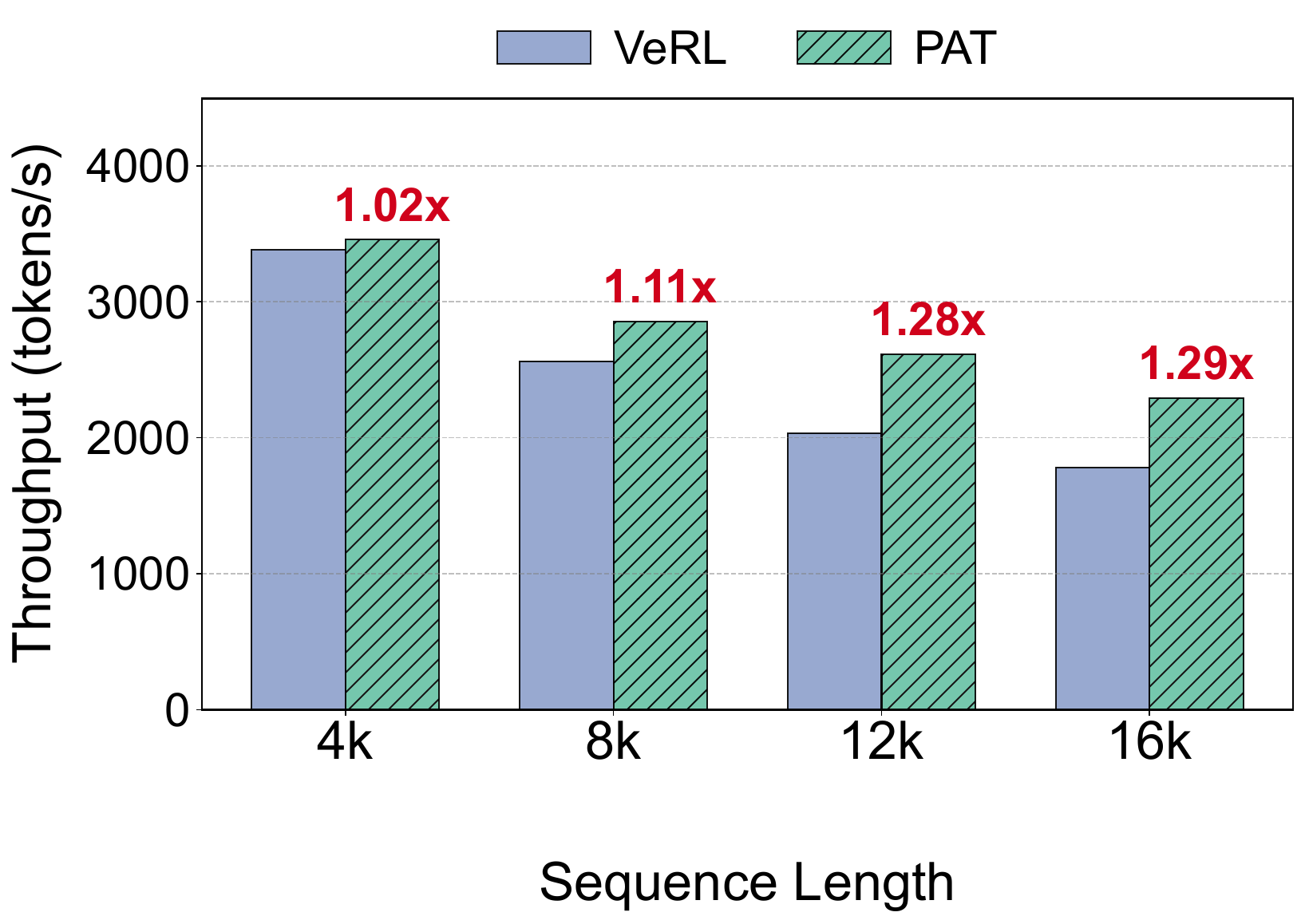}\label{fig:h100_14b_e2e}}
    \hspace{0.018\textwidth}
    \subfloat[Step-level latency breakdown]
    {\includegraphics[width=0.62\columnwidth]{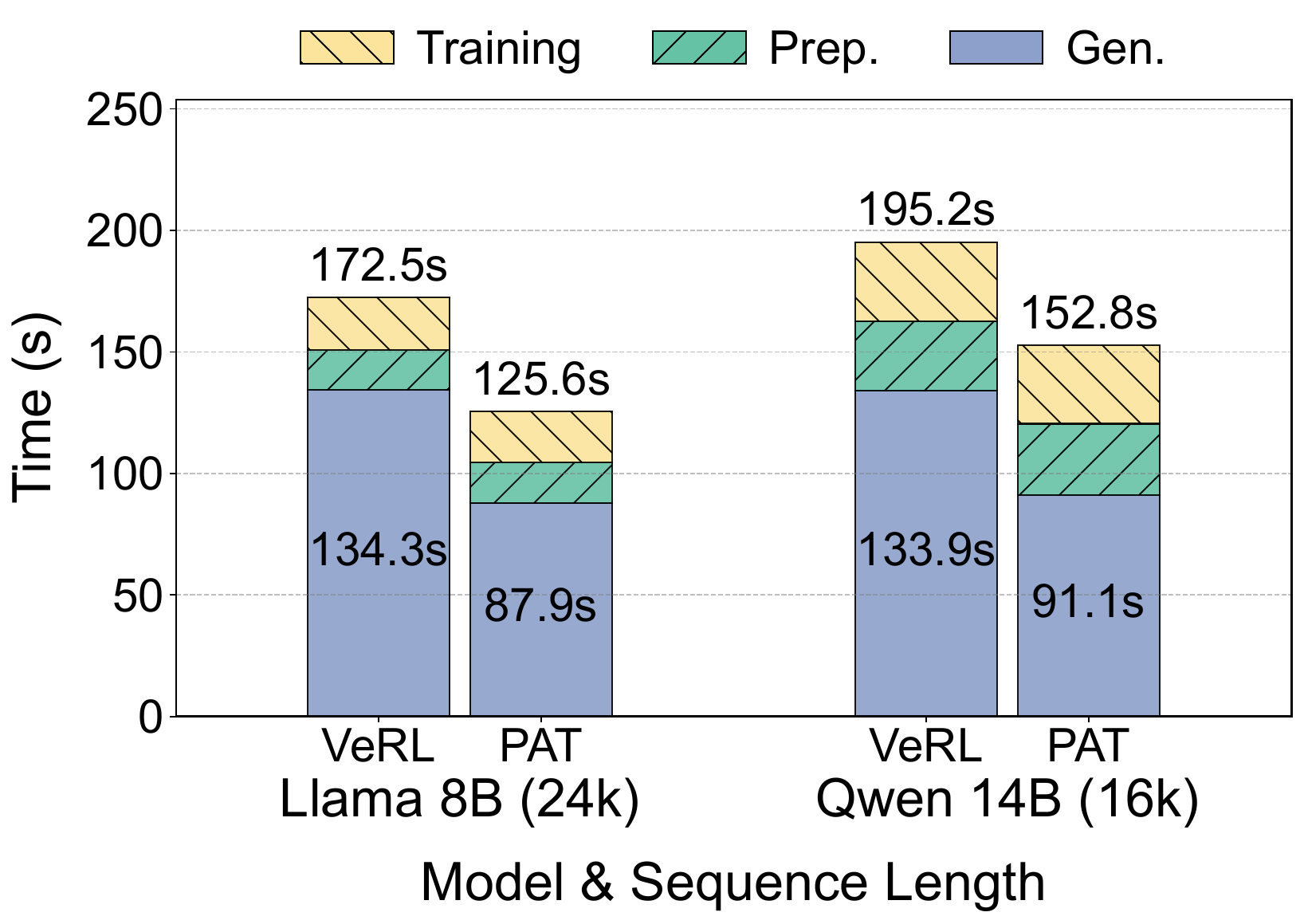}\label{fig:h100_breakdown}}
  }
  \caption{Comprehensive performance evaluation on the H100 server. (a) and (b) compare the end-to-end throughput under increasing sequence lengths for LLaMA3.1-8B and Qwen3-14B models. (c) illustrates the detailed time breakdown of training steps for LLaMA3.1-8B (24k) and Qwen3-14B (16k).}
  \label{fig:h100_comparison}
\end{figure*}

\section{Evaluation}
\subsection{Experimental Setup}
\para{Testbed.}
To comprehensively evaluate the performance and adaptability of \our{}, we conduct experiments on two types of GPU clusters. 
First, we deploy \our{} on a cluster of two nodes (16 GPUs total), with each node equipped with 8 NVIDIA A40 48GB GPUs connected via PCIe 4.0 $\times$ 16, while inter-node connectivity is provided by 100Gbps InfiniBand. 
Additionally, we evaluate performance on a single high-performance server equipped with 8 NVIDIA H100 80GB SXM GPUs, connected via NVLink with a bidirectional bandwidth of 900 GB/s.
Our experiments use the following software versions: CUDA 12.6, PyTorch 2.7.1, VeRL 0.4.1, and SGLang 0.4.8.

\para{Model and dataset.}
We adopt the widely used GRPO algorithm~\cite{shao2024deepseekmath} for RLHF. 
We use LLaMA3.1-8B~\cite{dubey2024llama} and Qwen3-14B~\cite{yang2025qwen3} as actor models. 
For the training data, we employ the DeepScaleR~\cite{luo2025deepscaler}, which is a logic-intensive dataset designed to enhance reasoning capabilities through RL. This dataset provides a representative stress test for \our{}, as its response length distribution exposes the tail-phase bottleneck targeted by our design.

\para{Baseline and metric.}
We compare \our{} against VeRL, a state-of-the-art colocated RLHF training framework. To ensure a fair comparison, for each model, hardware setting, and maximum response length, we tune the static TP/DP configuration of VeRL among feasible settings and report the best-performing baseline. Our primary evaluation metric is throughput, defined as the ratio of the total number of tokens generated in a single RLHF step to the duration of that step.
For each maximum response length, we report the average latency over five stable RLHF steps in which at least one generated sample reaches the specified maximum length.
% To evaluate the efficiency of \our{} on long-sequence workloads, we vary the maximum response lengths across different settings and compute the average over five stable steps where the sequence reaches its maximum length.

% After the warmup phase, we compute the mean throughput over 5 iterations. 
% To stress the system under peak workload and verify the efficiency of adaptive TP for long-sequence tasks, we vary the maximum response lengths across different experimental settings.

\subsection{End-to-End Throughput}

% \begin{figure}[!t]
%   \centering
%   \subfloat[Throughput under varying maximum response lengths]
%   {\includegraphics[width=0.48\columnwidth]{fig/llama_throughput.pdf}\label{fg6.1a}}
%   \hfill
%   \subfloat[Step-level breakdown in a representative iteration]
%   {\includegraphics[width=0.48\columnwidth]{fig/step_breakdown.pdf}\label{fg6.1b}}
% \caption{Performance comparison of \our{} and VeRL on the A40 cluster.}
% \label{fig:a40_performance}
% \end{figure}

% \begin{figure*}[!t]
%   \centering
%   \subfloat[Throughput of LLaMA3-8B on single-node (8$\times$A40)]
%   {\includegraphics[width=0.96\columnwidth]{fig/a40_8b_e2e_throughput.pdf}\label{fig:a40_8b_single}}
%   \hfill
%   \subfloat[Throughput of LLaMA3-8B on dual-node (16$\times$A40)]
%   {\includegraphics[width=0.96\columnwidth]{fig/a40_8b_e2e_multi_throughput.pdf}\label{fig:a40_8b_dual}}
%   \caption{End-to-end throughput comparison between VeRL and \our{} on the A40 cluster across different node configurations and maximum response lengths.}
%   \label{fig:a40_comparison}
% \end{figure*}
% We evaluate the end-to-end throughput of \our{} against VeRL across varying maximum response lengths, model architectures, and hardware configurations.

\para{Results on A40 Cluster.} We conduct end-to-end throughput evaluations on an A40 cluster to verify the effectiveness of \our{}, utilizing LLaMA3.1-8B and Qwen3-14B across both single-node (8$\times$A40) and dual-node (16$\times$A40) configurations with varying sequence lengths.

On the single 8$\times$A40 server with LLaMA3.1-8B (Fig.~\ref{fig:a40_8b_single}), \our{} achieves speedups ranging from 1.05$\times$ to 1.27$\times$. A notable trend is that the acceleration becomes increasingly pronounced as the sequence length increases (from 6k to 16k). This advantage stems from \our{}'s adaptive TP strategy. As the context length grows, the tail phase becomes more severe and occupies a larger proportion of the total runtime, thereby maximizing the benefits of our targeted optimization under preferred parallel configurations. This confirms the effectiveness of \our{} in mitigating severe tail bottlenecks.

When scaling the same LLaMA3.1-8B setup to the dual-node configuration (16$\times$A40, Fig.~\ref{fig:a40_8b_dual}), \our{} still delivers consistent improvements, achieving up to 1.19$\times$ speedup. 
Although the speedup is slightly limited by inter-node communication for DP synchronization during the training stage of each RLHF iteration, the consistent gains validate the scalability of \our{} in distributed settings.
% Although the overall acceleration is slightly tempered by the inter-node communication of DP synchronization in training stage, the results validate the scalability and efficiency of our adaptive design across distributed environments.

To verify the impact of different model architectures, we replace the model with the larger Qwen3-14B in the dual-node environment (Fig.~\ref{fig:a40_14b_dual}). Here, \our{} delivers speedups ranging from 1.05$\times$ to 1.18$\times$. We observe that these gains are relatively lower than those of LLaMA3.1-8B under the same dual-node setting. This disparity occurs because Qwen3-14B's larger communication volume increases the fraction of time spent on communication under the limited PCIe bandwidth, making it the dominant bottleneck. Nevertheless, consistent gains across model architectures confirm that \our{} remains robust and effective even under hardware bandwidth constraints.

\para{Results on H100 Server.}
To further validate the effectiveness of \our{} on high-end GPU platforms, we conduct an end-to-end throughput evaluation on a single 8-GPU H100 server. As shown in Fig.~\ref{fig:h100_8b_e2e} and Fig.~\ref{fig:h100_14b_e2e}, \our{} outperforms VeRL on both models. The H100 results mirror the trend observed on the A40 cluster: the relative speedup of \our{} increases with the maximum response length. For LLaMA3.1-8B, \our{} achieves 3510.9 tokens/s at a maximum response length of 24K, corresponding to a 1.37$\times$ speedup over VeRL (2570.6 tokens/s). For Qwen3-14B, \our{} achieves 2292.6 tokens/s at 16K, yielding a 1.29$\times$ speedup over VeRL (1780.0 tokens/s).

Notably, \our{} achieves larger speedups on the H100 server than on the A40 cluster. NVLink’s higher communication bandwidth reduces both the overhead of larger TP degrees and the cost of online reconfiguration. As a result, in the tail phase where only a few long responses remain active, higher TP more effectively reduces per-token latency and more easily amortizes the switching cost. This shows that \our{} can better exploit high-bandwidth GPU platforms for long-tail RLHF generation.
% Notably, the speedup on the H100 server is higher than that on the A40 cluster. One reason is that H100's higher communication bandwidth of NVLink reduces the relative overhead of using larger TP degrees. As a result, in the tail phase where only a few long responses remain active, switching to a larger TP degree brings a more pronounced per-token latency reduction than on A40. In addition, the lower communication cost also reduces the overhead of online reconfiguration, making the switch easier to amortize. \our{} therefore achieves larger tail-phase acceleration on H100, demonstrating that adaptive TP reconfiguration can better exploit high-end GPU platforms for long-tail RLHF generation.

\subsection{Step-level Breakdown}
To understand the source of the end-to-end speedup, we analyze the step-level latency breakdown of representative H100 runs. As shown in Fig.~\ref{fig:h100_breakdown}, \our{} mainly reduces the generation time, which is the stage targeted by adaptive TP reconfiguration.

For LLaMA3.1-8B with a maximum response length of 24K, \our{} reduces generation time from 134.3~s to 87.9~s, achieving a 34.6\% reduction. This leads to a 1.37$\times$ end-to-end speedup, with the overall step latency decreasing from 172.5~s to 125.6~s. For Qwen3-14B at 16K, \our{} similarly reduces generation time from 133.9~s to 91.1~s, corresponding to a 32.0\% reduction. Meanwhile, the preparation and training stages vary by less than 1.5\%, indicating that \our{} improves RLHF efficiency by directly reducing the long-tail generation bottleneck without shifting overhead to other stages.

\begin{figure}[!t]
\centerline{\includegraphics[width=0.9\linewidth]{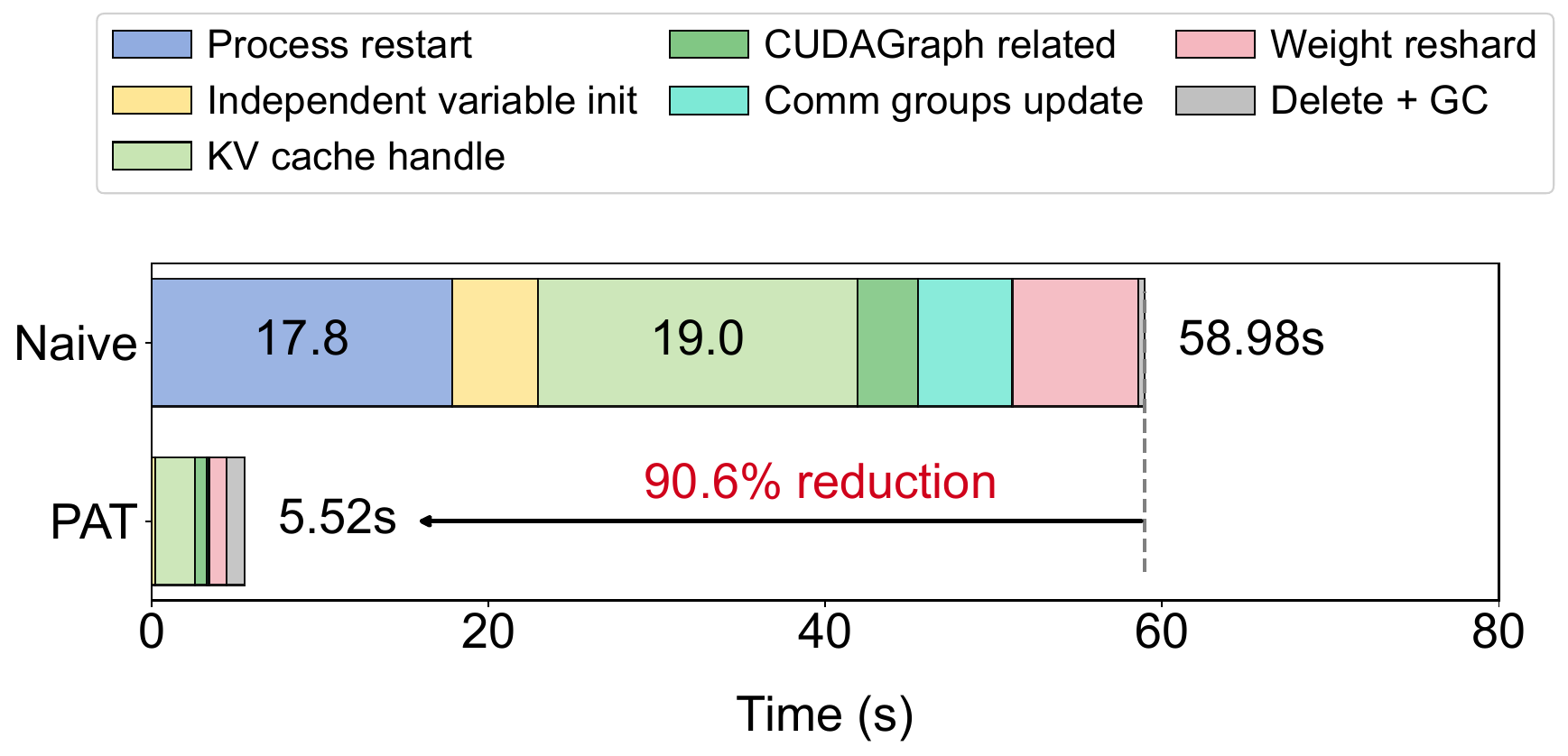}}
\caption{Switching latency breakdown comparison between Naive and \our{} on the A40 cluster (LLaMA3.1-8B, triggered at 12k context with 9 samples remaining).}
\label{switch_overhead}
\end{figure}

\subsection{Switching Overhead}

\para{Analysis of switching time.}
We evaluate the switching overhead of \our{} against a naive restart-based approach. We use the A40 cluster with a TP2-to-TP8 switch at a 12k context length and 9 remaining samples as a representative high-overhead case, due to PCIe bandwidth limits and large KV-transfer volume. As shown in Fig.~\ref{switch_overhead}, the naive approach takes 58.98~s due to process restarts, memory re-initialization, full KV recomputation, and weight reloading. In contrast, \our{} reduces the switching cost to 5.52~s, achieving a 90.6\% reduction. This overhead accounts for only 1.23\% of the 447~s end-to-end runtime. Without these optimizations, the runtime would increase by 53.46~s to 500.46~s, offsetting much of the tail-phase acceleration from larger TP.

The reduction in switching overhead mainly comes from optimized data movement and runtime-state reuse. First, reload-free layer-wise redistribution reduces weight resharding from 7.47~s to 1.03~s. Second, migrating only valid KV states reduces KV handling from 19.01~s to 2.36~s compared with full prefill. Third, CUDA Graph recapturing is reduced from 3.59~s to 0.73~s by only targeting the small batch sizes that may appear after switching, while communication groups and other independent initialization tasks are reused when possible. We also observe that the switching overhead remains nearly unchanged in multi-node settings under the same 12K-token context length at the switching point, because TP switching is confined to each node, enabling parallel switching across independent DP groups. Thus, the overall latency is determined by the per-node switching time rather than the number of nodes.

\begin{figure}[!t]
\centerline{\includegraphics[width=0.9\linewidth]{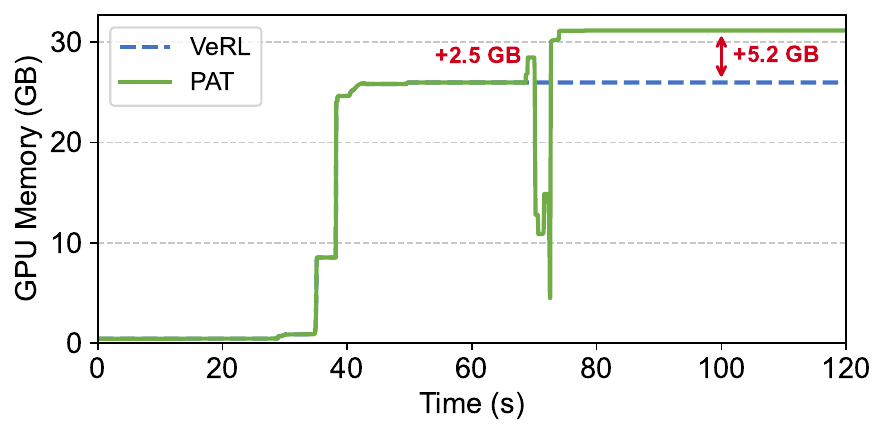}}
\caption{Memory footprint during the switching phase for LLaMA3.1-8B on the DeepScaleR dataset using 8$\times$A40 single-node configuration.}
\label{memory_usage}
\end{figure}

\para{Analysis of extra memory overhead.}
We further analyze the memory overhead during switching. As shown in Fig.~\ref{memory_usage}, a naive full-state migration would require moving the entire 14~GB KV cache and may trigger OOM. In contrast, \our{} increases peak memory usage by only 2.5~GB during switching, because it extracts only valid KV payloads and redistributes them layer by layer. The 5.2~GB increase in steady-state memory after switching comes from reallocating a larger KV cache pool required by the new TP configuration, rather than from intrinsic overhead introduced by \our{}.

\subsection{Predictor Effectiveness}
\label{subsec:predictor_effectiveness}
To validate the reliability of our reconfiguration decisions, we evaluate whether the predictor built from dummy-batch profiling can accurately estimate performance on real dataset-derived workloads. Although the \texttt{Offline Profiler} collects latency profiles using synthetic fixed-shape batches rather than dataset samples, we compare the predicted throughput against ground-truth measurements on real decoding trajectories from the DeepScaleR~\cite{luo2025deepscaler} dataset under TP degrees of 2 and 8.

\begin{figure}[!t]
  \centering
  \begin{minipage}[b]{0.48\linewidth}
    \centering
    \includegraphics[width=\linewidth]{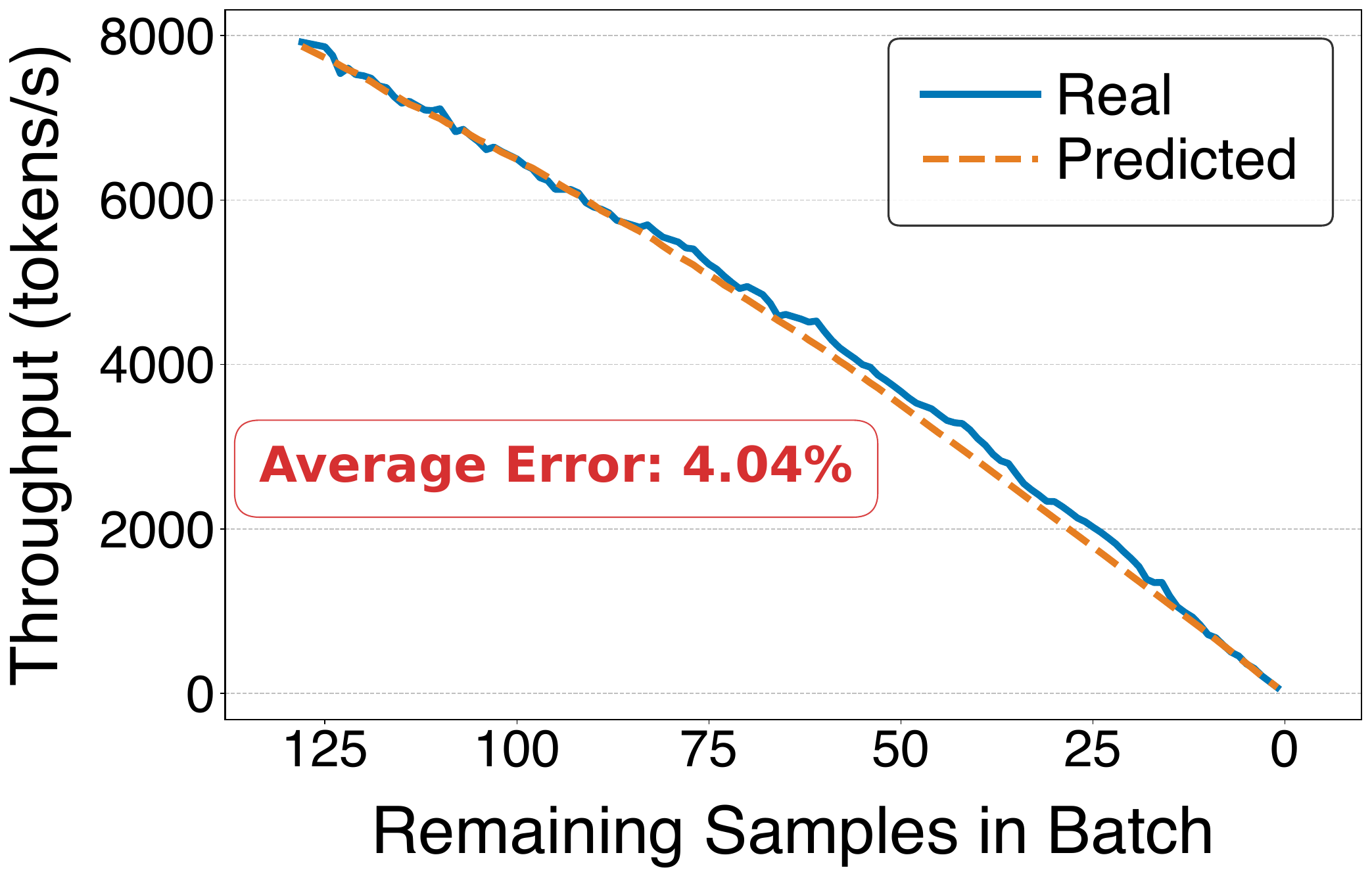}
    \centerline{\footnotesize (a) DeepScaleR (TP=2, DP=4)}
  \end{minipage}
  \hfill
  \begin{minipage}[b]{0.48\linewidth}
    \centering
    \includegraphics[width=\linewidth]{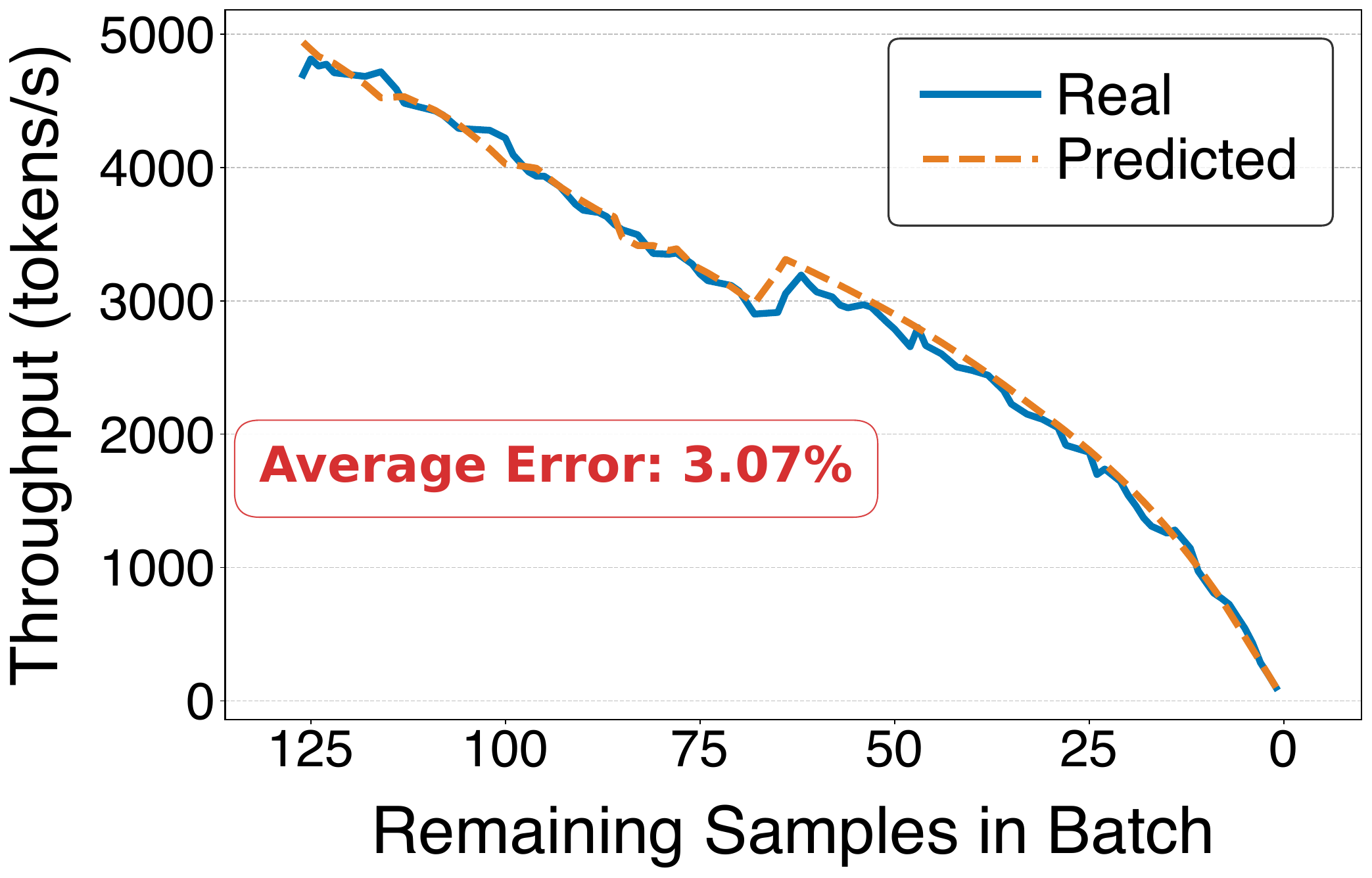}
    \centerline{\footnotesize (b) DeepScaleR (TP=8, DP=1)}
  \end{minipage}

  % \vspace{0.5em}

  % \begin{minipage}[b]{0.48\linewidth}
  %   \centering
  %   \includegraphics[width=\linewidth]{fig/hhrlhf-dp4tp2.pdf}
  %   \centerline{\footnotesize (c)
  %   HH-RLHF (TP=2)}
  % \end{minipage}
  % \hfill
  % \begin{minipage}[b]{0.48\linewidth}
  %   \centering
  %   \includegraphics[width=\linewidth]{fig/hhrlhf-dp1tp8.pdf}
  %   \centerline{\footnotesize (d) HH-RLHF (TP=8)}
  % \end{minipage}

  \caption{Evaluation of throughput prediction accuracy across different TP degrees. The plots compare the real and predicted throughput on the DeepScaleR dataset.}
  \label{fig:predictor_acc}
\end{figure}

As illustrated in Fig.~\ref{fig:predictor_acc}, the predictor closely matches the real throughput trajectories as the number of remaining samples decreases during batch draining. Quantitatively, it achieves low average prediction errors of 4.04\% for TP=2, DP=4 and 3.07\% for TP=8, DP=1 on the DeepScaleR dataset. Notably, as shown in Fig.~\ref{fig:predictor_acc}b, the TP8 trajectory exhibits a local non-monotonic fluctuation caused by shape-dependent backend behavior. Despite this irregularity, the predictor captures both the overall decreasing trend and the local throughput jump. These results demonstrate that our profiling method is dataset-independent and can still provide high-accuracy predictions for real decoding workloads.

\subsection{Discussion}
Long-tail generation is common in synchronous RLHF workloads rather than a corner case. Prior work has observed pronounced response-length skew in open-ended dialogue traces such as LMSYS-Chat-1M, as well as in reasoning-oriented math/code RL workloads~\cite{zhong2025optimizing,zhong2025streamrl}. \our{} targets this common setting by adapting the TP/DP configuration as the active batch size shrinks during decoding. For workloads with short or nearly uniform responses, the tail phase is less pronounced, so \our{} may provide smaller gains; in such cases, the predictor avoids switching unless the estimated benefit exceeds the reconfiguration cost.

%% file: 6.relatedwork.tex
\section{Related Work}

% \para{RL training frameworks.}
Several systems have been proposed to improve the efficiency and usability of RLHF training.
OpenRLHF~\cite{hu2024openrlhf} provides an easy-to-adopt open-source platform, but typically partitions cluster resources across different RLHF components, which can lead to low utilization during sequential stage execution.
VeRL~\cite{sheng2025hybridflow} improves resource sharing through a colocated architecture and flexible RLHF data flows, yet its generation stage still relies on static parallelism and remains vulnerable to long-tail decoding.
StreamRL~\cite{zhong2025streamrl} mitigates stage-level bubbles by enabling asynchronous execution between generation and training, but such benefits are limited in strictly synchronous RLHF settings and may introduce stale-rollout trade-offs.
\our{} instead focuses on improving generation efficiency within the synchronous execution model by adaptively reconfiguring the TP degree during a single generation stage.

% \para{Mitigating the generation long-tail problem.}
Recent systems have explored different ways to mitigate long-tail generation in RLHF.
RLHFuse~\cite{zhong2025optimizing} mitigates long-tail stalls through inter-stage overlap, hiding part of the tail latency with preparation-stage execution.
However, this benefit relies on sufficient preparation work and can be limited when the preparation stage is short, such as in GRPO without a critic, settings without a reward model, or iterations dominated by long-tail generation.
Moreover, inter-stage overlap does not directly reduce the decoding latency of the tail samples themselves.
Kimi K2~\cite{team2025kimi} adopts partial rollout, which pauses unfinished long-tail trajectories and resumes them in later RL iterations to prevent them from blocking the current rollout process.
However, this breaks synchronous RLHF semantics because different parts of the same trajectory may be generated by different actor checkpoints, introducing off-policy or mixed-policy rollouts and potentially posing risks to training stability or final accuracy.
In contrast, \our{} preserves synchronous RLHF semantics and directly accelerates tail-phase decoding within a single generation stage through adaptive TP reconfiguration.

% For example, DistServe\cite{zhong2024distserve}  decouples the Prefill and Decoding stages to reduce the interference of Prefill on Decoding while satisfying their respective service-level objectives (SLOs). LoongServe\cite{wu2024loongserve} improves generation efficiency by assigning different sequence parallelism degrees to requests of varying lengths. Notably, these methods are orthogonal to the optimizations of \our{}, and can be further integrated.

% Besides, some works~\cite{zhong2025streamrl, noukhovitch2024asynchronous} have deployed asynchronous RL to increase the batch size of the tail phase. 

\section{Conclusion}
% \noindent We present \our{}, an efficient and flexible RLHF training framework for LLMs. Motivated by an analysis of long-tail behaviors in the generation stage, as well as observations of decoding latency variations across different TP degrees, we introduce a runtime mechanism for dynamically reconfiguring TP degrees, which substantially improves generation efficiency. %%CL with solid 
We present PAT, an adaptive TP reconfiguration framework for accelerating long-tail generation in synchronous RLHF training. PAT dynamically adjusts the TP/DP configuration within each generation stage to better match the aligned and tail phases. Implemented on SGLang and integrated with VeRL, PAT reduces generation latency by up to 34.6\% and training iteration time by up to 27.2\%.